\documentclass[HARVARD,Times2COL]{WileyNJDv5}

\articletype{Research Article}

\usepackage[para,online,flushleft]{threeparttable}

\received{Date Month Year}
\revised{Date Month Year}
\accepted{Date Month Year}
\journal{J Field Robotics}
\volume{00}
\copyyear{2024}
\startpage{1}

\begin{document}
\title{Boundary Control Behaviors of Multiple Low-cost AUVs Using Acoustic Communication}

\author[1]{Mohammed Tarnini}

\author[1]{Saverio Iacoponi}

\author[1]{Andrea Infanti}

\author[2]{Cesare Stefanini}

\author[1]{Giulia De Masi}

\author[1]{Federico Renda}

\authormark{Tarnini \textsc{et al.}}
\titlemark{Boundary Control Behaviors of Multiple Low-cost AUVs Using Acoustic Communication}

\address[1]{\orgdiv{Department of Mechanical and Nuclear Engineering}, \orgname{Khalifa University}, \orgaddress{\state{Abu Dhabi}, \country{United Arab Emirates}}}

\address[2]{\orgdiv{The BioRobotics Institute}, \orgname{Scuola Superiore Sant'Anna}, \orgaddress{\state{Pisa}, \country{Italy}}}

\corres{Mohammed Tarnini, Mechanical and Nuclear Engineering Department, Khalifa University, Abu Dhabi, United Arab Emirates. \email{mohammedtarnini@gmail.com}}

\fundingInfo{Technology Innovation Institute Fund (contract no. TII/ARRC/2047/2020)}

\abstract[Abstract]{
This study presents acoustic-based methods for the control of multiple autonomous underwater vehicles (AUV). This study proposes two different models for implementing boundary and path control on low-cost AUVs using acoustic communication and a single central acoustic beacon. Two methods are presented: the Range Variation-Based (RVB) model completely relies on range data obtained by acoustic modems, whereas the Heading Estimation-Based (HEB) model uses ranges and range rates to estimate the position of the central boundary beacon and perform assigned behaviors. The models are tested on two boundary control behaviors: Fencing and Milling. Fencing behavior ensures AUVs return within predefined boundaries, while Milling enables the AUVs to move cyclically on a predefined path around the beacon. Models are validated by successfully performing the boundary control behaviors in simulations, pool tests, including artificial underwater currents, and field tests conducted in the ocean. All tests were performed with fully autonomous platforms, and no external input or sensor was provided to the AUVs during validation. Quantitative and qualitative analyses are presented in the study, focusing on the effect and application of a multi-robot system.
}


\maketitle
\section{Introduction}\label{Introduction}
Autonomous Underwater Vehicles (AUVs) are self-operating robots essential for exploring and mapping the largely unexplored ocean and ocean floor, often working together to achieve shared goals \cite{Yang2021}. The potential applications of AUVs are vast, where they can be used in environmental monitoring and collecting data on physical, chemical, and biological parameters \cite{Das2016}. Exploring or monitoring vast expanses of seabed and open water presents significant challenges for individual vehicles. Unlike aerial and terrestrial domains, which can utilize aerial drones or satellites to cover large areas from considerable distances, underwater environments impose significant limitations due to restricted visibility, limited communications, and inability to use GPS or other established wide-area localization systems \cite{Cardenas2024}. Consequently, the detection range of an AUV is severely constrained. Moreover, environmental factors such as waves and underwater currents add further challenges to AUVs' deployment and navigation \cite{Peng2019}. Therefore, a multi-AUV approach becomes strongly advantageous to enable task parallelization and introduce redundancy, thereby mitigating the risk of mission failure in the event of individual AUV malfunction. Within the domain of multi-AUV systems, Swarm Robotics emerges as a promising field in facing those challenges \cite {Luvisutto2022}. Comprising small, agile, and autonomous units, swarms possess the capacity for intra-swarm communication, facilitating information exchange and collaborative decision-making among the AUVs. Swarm Robotics draws inspiration from their natural counterparts, such as the schools of fish, flocks of birds, and swarms of bees. While swarm robotics has been extensively studied in terrestrial and aerial contexts \cite{Ricardo2023}, \cite{Yin2023}, \cite{Nisser2022}, and \cite{Starks2023}  examples of underwater swarm robots remain relatively scarce.

Innovative methods in the multiple AUVs field have been implemented to enhance navigation, coordination, and information collection \cite{Wang2023}, \cite{Wang2022}. For example, Salazar et al. used Soft Robotic Fish (SoFi) to perform a leader-follower control method \cite{Salazar2022}. Wang et al. also implemented the leader-follower method with the addition of potential fields for controlling multiple AUVs in a dynamic formation, prioritizing obstacle avoidance, and efficient communication \cite{Wang2023}. Regarding underwater navigation, \cite{Matsuda2022} built upon previous works \cite{Matsudea2012}, \cite{Matsuda2018}, and \cite{Matsuda2019} and developed a method for multiple AUV navigation, where inexpensive AUVs accurately determine their positions, orientations, and velocities using acoustic communication with one high-performance AUV. \cite{Wu2019} focused on maximizing information acquisition by developing strategies for recording videos with multiple AUVs, facilitating the reconstruction of underwater archaeological sites. Additionally, \cite{Pi2021} proposed a solution for transferring long underwater tubes using multiple AUVs with the aim of reducing the operation expenses associated with Remotely Operated Vehicles (ROVs). A decentralized Task-Priority kinematic control algorithm was utilized and tailored for restricted communication capacity involving two 8 Degree of Freedom (DoF) AUVs.

One of the few experimental applications of the swarm robotics paradigm in the underwater domain is found in the BlueBot project \cite{Berlinger2021}. BlueBot is comprised of a swarm of fish-inspired robots engineered to emulate the collective behaviors observed in schools of fish. Noteworthy achievements of the project include the realization of behaviors such as dispersion, aggregation, and Milling. Collaboration among individual robots is facilitated through the implementation of elementary shared rules, as well as visual feedback. Experimental validation was conducted within a dark and controlled environment. Inspired by the adaptive nature of honeybees, \cite{Zahadat2016} applied a decentralized algorithm for flexible task allocation and adaptive partitioning on a swarm of AUVs. The algorithm dynamically adjusts to changes occurring in swarm size and task demands, which results in swarm behaviors' successful implementation under varying conditions without requiring global information. Collaboration between AUVs could also be used for precise data acquisition for an underwater environment with rapidly changing conditions. Meyer et al. utilized a swarm of small AUVs designed with environmental sensors to monitor oceanic submesoscale eddies \cite{Meyer2017}. The AUVs obtained water column measurements on diverse paths using acoustic communication and saw tooth diving behavior. Similarly, \cite{Ivan2019} introduced an underwater acoustic sensor network using a heterogeneous robotic swarm for extended underwater environment monitoring, highlighting interactions, hardware capabilities, and communication protocols.

More instances of swarm robotics applications can be found within the realm of simulation. \cite{Bodi2015}  introduces BEELCLUST, a bio-inspired algorithm tailored for exploration tasks, drawing inspiration from the heat-seeking behavior of bee swarms. \cite{Jia2019} delve into the utilization of swarm robotics for search and exploration missions targeting moving entities, leveraging simulated visual and acoustic inputs. \cite{Cheah2004} demonstrate methods for boundary control behavior, delineating target spatial regions and employing potential field techniques to guide swarms of AUVs within these regions. Subsequent studies by the same team extend these methods to accommodate collision avoidance among AUVs, dynamic target regions \cite{Cheah2008}, and navigation amidst external obstructions and obstacles \cite{Cheah2009}, \cite{Hou2011}. A similar strategy is presented by \cite{Ismail2011}, \cite{Ismail2012}, wherein the target area is defined as the boundary of a spatial region; this method was of particular interest for pipe following and inspection algorithms. \cite{Haghighi2012} exemplify multi-group coordination control, employing a combination of simple shape boundaries to orchestrate complex formations. Notably, these aforementioned studies assume complete knowledge of the AUVs' states. Regarding the control of multi-robot systems with limited state information, \cite{Wang2013} and \cite{Wang2021} demonstrate Milling behavior in simulation scenarios despite possessing only partial information regarding the AUVs' positions relative to a central point (range) and the relative locations of their nearest neighbors.

Several studies have been conducted for AUVs and underwater glider navigation using a single underwater beacon \cite{Wang2022_1}, \cite{Yuan2022}, \cite{Liu2020}. These studies capitalize on obtaining state estimation of AUVs through dead reckoning and subsequently refining the estimations using data received from the single beacon. In our work, we rely entirely on communication with the single beacon for controlling the AUV, without onboard state estimation. Moreover, it is evident that the majority of research pertaining to the control of underwater robotic swarms has been predominantly conducted within simulation environments, with limited examples of real-world experimentation. In contrast, the realm of multi-UAV systems has received more rigorous investigation \cite{Dorigo2021}. Nonetheless, there exists a noticeable gap in the availability of control algorithms for multi-AUV systems that are both easily scalable and conceptually straightforward without relying on complete and readily available estimations of each AUV's state. Based on an existing set of AUVs developed under the scope of the H-SURF project \cite{Iacoponi2022}, a suite of behavioral algorithms for multi-AUV control is proposed. Specifically, we introduce behaviors such as spatial confinement (Fencing) and Milling, which eschew reliance on onboard localization in favor of leveraging acoustic ranging and Doppler shift in communication. This approach is of particular interest for scalability with increasing numbers of AUVs, thereby facilitating its implementation within large-scale multi-AUV and swarm systems. The scalability can be achieved by leveraging the utilization of broadcasts and Doppler shift, limiting the number of communication packages required as the number of AUVs increases.

The paper is organized in the following manner. An explanation of the methods is shown in Section \ref{Method}. Section \ref{H-SURF AUV} presents the platform used in experiments: the H-SURF system.  Pool experiments are then shown in Section \ref{Pool Experiments}, and field experiments are shown in \ref{Field Experiments}. Discussion and comparison between results is presented in Section \ref{Discussion}. Finally, the conclusions of the main results are presented in Section \ref{Conclusions}.

\section{Method}
\label{Method}
\subsection{Overview}
In this section, we present several control algorithms implemented on our platforms. The control feedback is based on the relative distance from the central beacon, referred to as ranging and heading information. The first one is measured by the acoustic modem, while the second can be inferred by the onboard IMU and magnetometer data. Depth control ensures that all AUVs maintain approximately the same altitude, thereby simplifying the system to a two-dimensional problem. Horizontal control of the AUVs is characterized by under-actuation with two degrees of freedom: surge and yaw angle. While surge and heading can be controlled independently, sway is inherently uncontrollable. Rotations on place are accounted for, so the state space ($x,y,\psi$) is fully reachable. The model assumes control inputs in the form of surge force ($f_x$) and heading direction ($\psi$). It is important to note that the presented method operates under the assumption that AUVs exhibit sufficiently good control over these two parameters. 

\begin{figure}[!hbt]
    \centering
    \includegraphics[width=0.4\textwidth]{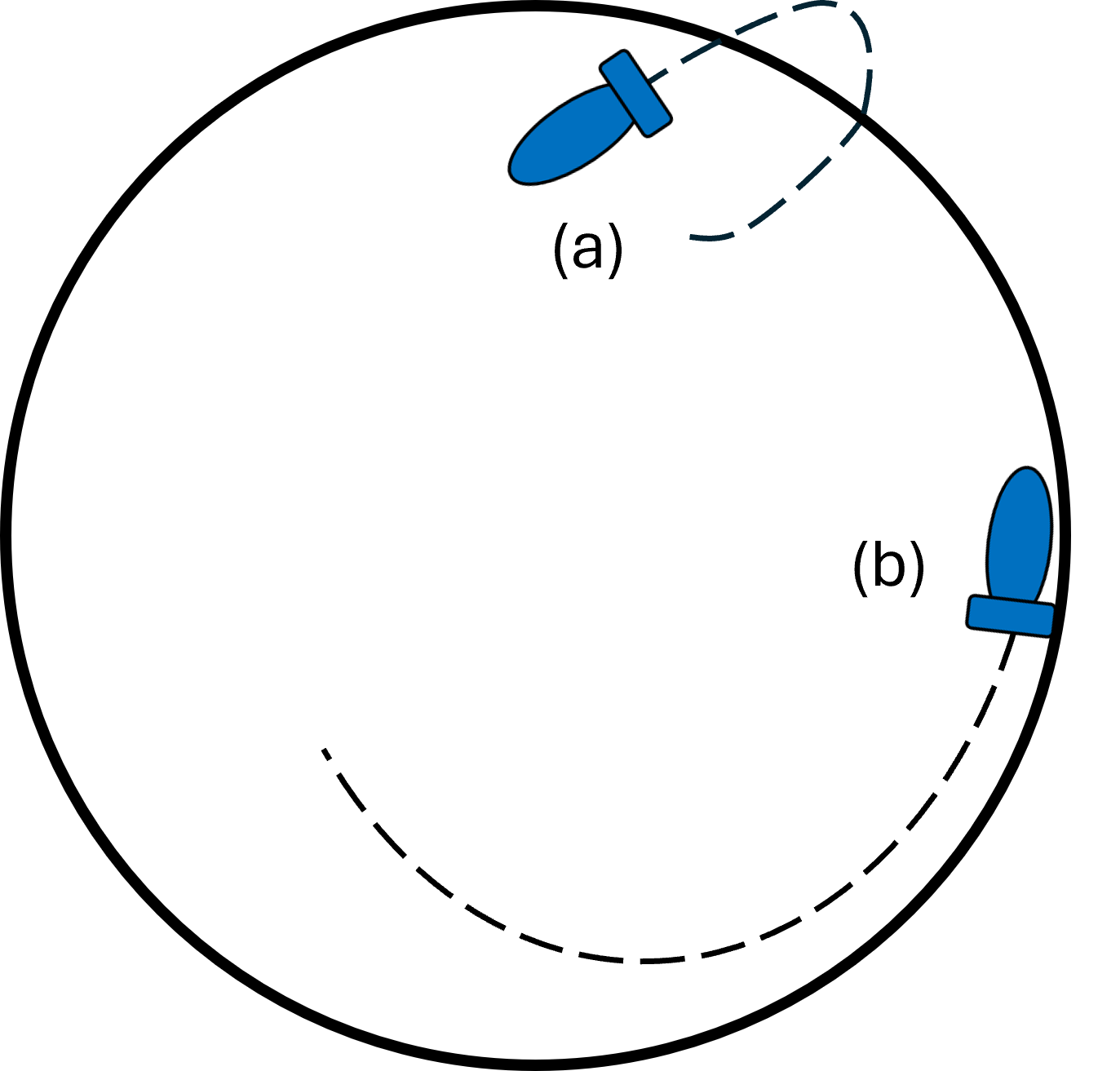}
    \caption{Overview of behaviors. (a) Fencing: the AUV is free to move within the boundary, but when it exceeds the boundary, the algorithm guides the AUV back towards the center. (b) Milling: The algorithms guide the AUV on a closed loop path around the center.}
    \label{Figure_1}
\end{figure}

The control algorithms serve to achieve two distinct behaviors, referred to as Fencing and Milling. The Fencing behavior aims to confine AUVs within predefined boundaries and ensures their prompt return if they stray beyond the defined boundaries, as illustrated in Figure \ref{Figure_1} (a). Notably, this behavior becomes active only when an AUV surpasses the designated boundaries, allowing AUVs to exhibit other desired behaviors while within the boundaries. On the other hand, the goal of the Milling behavior is to guide the AUVs along a close planar path centered around a point. Provided a closed path containing the center point and a specified rotation direction, the control algorithm should enable the AUVs to converge onto the path and traverse it in the indicated direction. Figure \ref{Figure_1} (b) provides a visual depiction of this behavior, which mimics the Milling patterns observed in natural schools of fish.

The proposed algorithms operate under the assumption of constant depth, with minor deviations in depth between the AUV and the central beacon compared to the horizontal distances under consideration. Additionally, these algorithms presume that all AUVs move at a constant surge velocity. While this assumption holds for the proposed methodology, empirical evidence suggests its successful application in more generalized scenarios, as demonstrated in the experimental analysis.

For both behaviors, two distinct approaches are proposed: Range Variation-Based (RVB) and Heading Estimation-Based (HEB). The RVB algorithms leverage the last two or three range measurements and employ straightforward rules to achieve the desired behaviors. However, the simplicity of this method restricts the options available for defining boundaries and paths. In contrast, the HEB algorithm takes in the historical series of rate-of-change in ranges in order to estimate the beacon's relative direction and, along with the most recent range measurement, enables the implementation of more intricate paths and boundaries.

\subsection{RVB Model}
\label{subsec:RVB}
\begin{figure}[!hbt]
    \centering
    \includegraphics[width=0.4\textwidth]{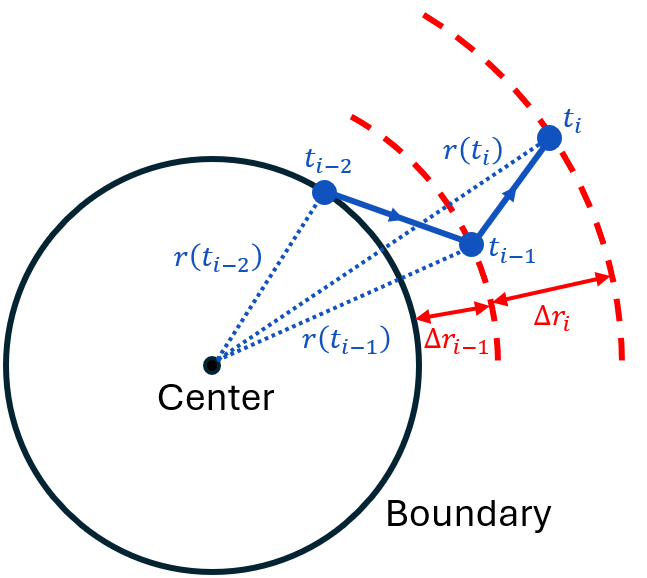}
    \caption{RVB Fencing ranges representation. Shown AUV position at times $t_i,t_{i-1},t_{i-2}$. As the AUV leaves the boundary, the last three ranges are recorded ($r(t_{i}),r(t_{i-1}),r(t_{i-2})$), and the range variation calculated ($\Delta r_i, \Delta r_{i-1}$). }
    \label{Figure_2}
\end{figure}

In RVB, there is exclusive reliance on range information for executing Fencing and Milling behaviors. RVB Fencing behavior requires AUVs to use the last three ranges $r(t_i), r(t_{i-1})$, and $r(t_{i-2})$ relative to the center of the bounded area. $\Delta r_i$ represents the last range increment, while $\Delta r_{i-1}$ denotes the previous range increment. The illustration seen in Figure \ref{Figure_2} visualizes the representation of ranges and range increments. 

Upon exiting the boundary, the Fencing motion is executed. Initially, the RVB Fencing algorithm selects an arbitrary direction of rotation and executes a rotation of a defined increment $\Delta \psi$. Then, it compares the last two range increments: if $\Delta r_{i} > \Delta r_{i-1}$, meaning that the AUV is rotating away from the boundary, the direction of rotation is inverted, otherwise it remains unchanged. When a new range is received, a new rotation increment is applied. 
Applying the aforementioned algorithm, RVB Fencing alters its rotation direction until eventually converging close to the ideal heading for the AUVs to return back to the boundary. A pseudo-code description of the RVB Fencing is shown in Algorithm \ref{Algorithm_1}. Since this method relies solely on range information and the relative direction of the central beacon is not directly estimated, only a circular boundary with a central beacon can be enforced. 

\begin{algorithm}
\caption{\enskip RVB Fencing Algorithm}
\begin{algorithmic}
\State \textbf{Event:} Receiving new $r(t_i)$
\State $\Delta r_{i-1} = r(t_{i-1}) - r(t_{i-2})$ 
\State $\Delta r_i = r(t_{i}) - r(t_{i-1})$ 
\If {$r(t_{i}) >$ circle radius}
\If {$\Delta r_i >$ $\Delta r_{i-1}$}
\State Change direction, $D = D \cdot -1$
\EndIf
\State heading = current heading + $\Delta \psi \cdot D$
\Else
\State Execute any desired in-boundary commands  
\EndIf
\end{algorithmic}
\label{Algorithm_1}
\end{algorithm}

\begin{figure}[!hbt]
    \centering
    \includegraphics[width=0.4\textwidth]{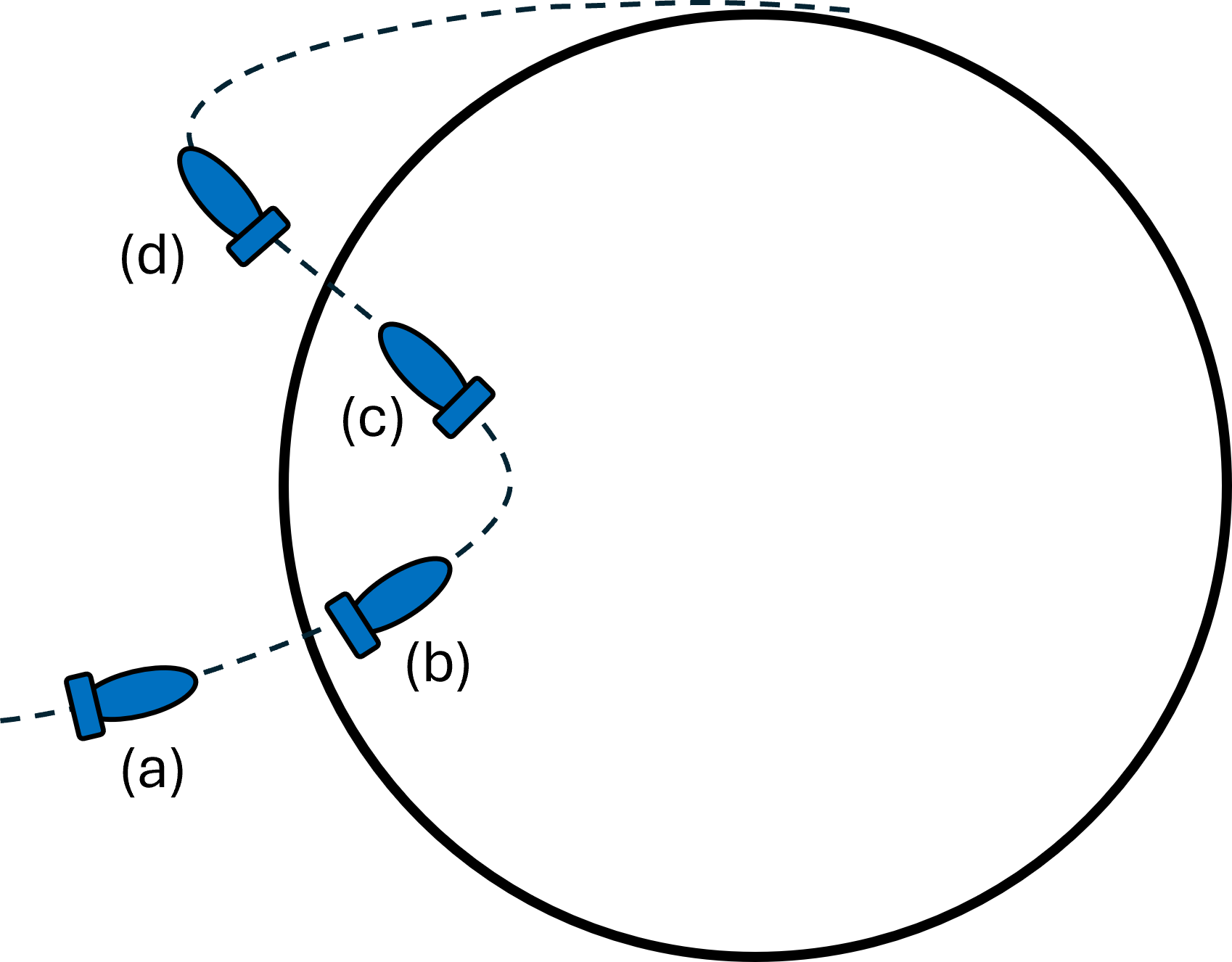}
    \caption{Possible conditions of AUV during Milling. Condition (a) represents the scenario where the AUV is beyond the boundary but approaching it. In condition (b), AUV is within the boundary and moving towards the center. Condition (c) depicts the AUV inside the boundary and moving away from the center. Condition (d) occurs when the AUV is outside the boundary and moving away.}
    \label{Figure_3}
\end{figure}

The application of the RVB algorithm to the Milling behavior takes the last two ranges and the last range increment as input. The desired rotation direction $D$ (Clock-Wise CW or Counter-Clock-Wise CCW) is set as the initial parameter.
The algorithm checks for four possible conditions in which the AUV can result in during the Milling behavior: (a) AUV outside the path moving closer to the center, (b) AUV inside the path moving closer, (c) AUV inside the path moving farther, (d) AUV outside the path moving farther. These conditions are illustrated in Figure \ref{Figure_3}. For conditions (a) and (c), the AUV keeps its current heading to move closer to the boundary. For conditions (b) and (d), the heading of the AUV is corrected according to the following equation:

\begin{equation}\label{Equation_1}
\Delta \psi_{cmd} = D \ k \ (r - R_0) 
\end{equation}

Where $k$ is a proportional parameter that determines the correction strength. The direction of rotation is determined by the sign of ($D=\pm 1$). $r$ is the range between AUV and the center of the boundary, and $R_0$ is the radius of the desired path.
Additionally, for small values of $R_0$, it is convenient, in practice, to add a periodic increment of the heading such that:

\begin{equation}
    \dot{\psi} = 
    \begin{cases}
        D \ k_{rate} \ \frac{R_0}{r} & \quad \textit{if} \ r \geqslant  R_0 \\ 
        0 & \quad \textit{if} \ r < R_0
    \end{cases}
    \label{Equation_2}
\end{equation}

Where $\dot{\psi}$ is a constant rotation applied to the AUVs, and $k_{rate}$ determines the strength of the constant rotation. The value of $k_{rate}$ is selected, given a constant value of surge velocity, such that the final motion of the AUV, excluding other corrections, would result in a circular motion with a radius significantly larger than $R_0$. This mitigates the communication losses, and while it is insufficient, on its own, to achieve milling, it contributes to strengthening the consistency of the milling. Thus, given the current measured heading $\psi_{IMU}$, the commanded heading is obtained with the following equation:

\begin{equation}
    \psi_{cmd} = \psi_{IMU} + \Delta\psi_{cmd}
    \label{Equation_3}
\end{equation}
Every $\Delta t$ seconds, the heading command is updated with:

\begin{equation}
    \psi_{cmd} \gets \psi_{cmd} + \dot{\psi} \ \Delta t 
    \label{Equation_4}
\end{equation}

Where $\Delta t$ is an arbitrarily short interval in between heading updates. The algorithm \ref{Algorithm_2} showcases the RVB Milling Algorithm.

\begin{algorithm}
\caption{\enskip RVB Milling Algorithm}\label{Algorithm_2}
\begin{algorithmic}
\State \textbf{Range Update Loop}
\State receiving new $r(t_{i})$
\State $\Delta r_i = r(t_{i}) - r(t_{i-1})$ 
\If {($r(t_{i}) < R_0$ and $\Delta r_i < 0$) or ($r(t_{i}) > R$ and $\Delta r_i > 0$)}
\State $\Delta \psi_{cmd} = D \ k \ (r(t_{i}) - R_0)$
\State $\psi_{cmd} = \psi_{IMU} + \Delta \psi_{cmd}$
\EndIf

\vspace{1em}

\State \textbf{Range Heading Loop}
\State timer 1 second
\If {$R_0$ is small and $r(t_{i}) > R_0$}
\State $\dot{\psi} = D \ k_{rate} \frac{R_0}{r(t_{i})} $
\State $\psi_{cmd} \gets \psi_{cmd} + \dot{\psi} \Delta t$
\EndIf
\end{algorithmic}
\end{algorithm}

The RVB algorithm for Milling is limited to circular paths with beacons located in the center. However, the algorithm's reliance on simple rules makes it more resilient to inaccuracies than algorithms requiring a full state estimation.

\subsection{HEB Model}
\begin{figure}[!hbt]
    \centering
    \includegraphics[width=0.4\textwidth]{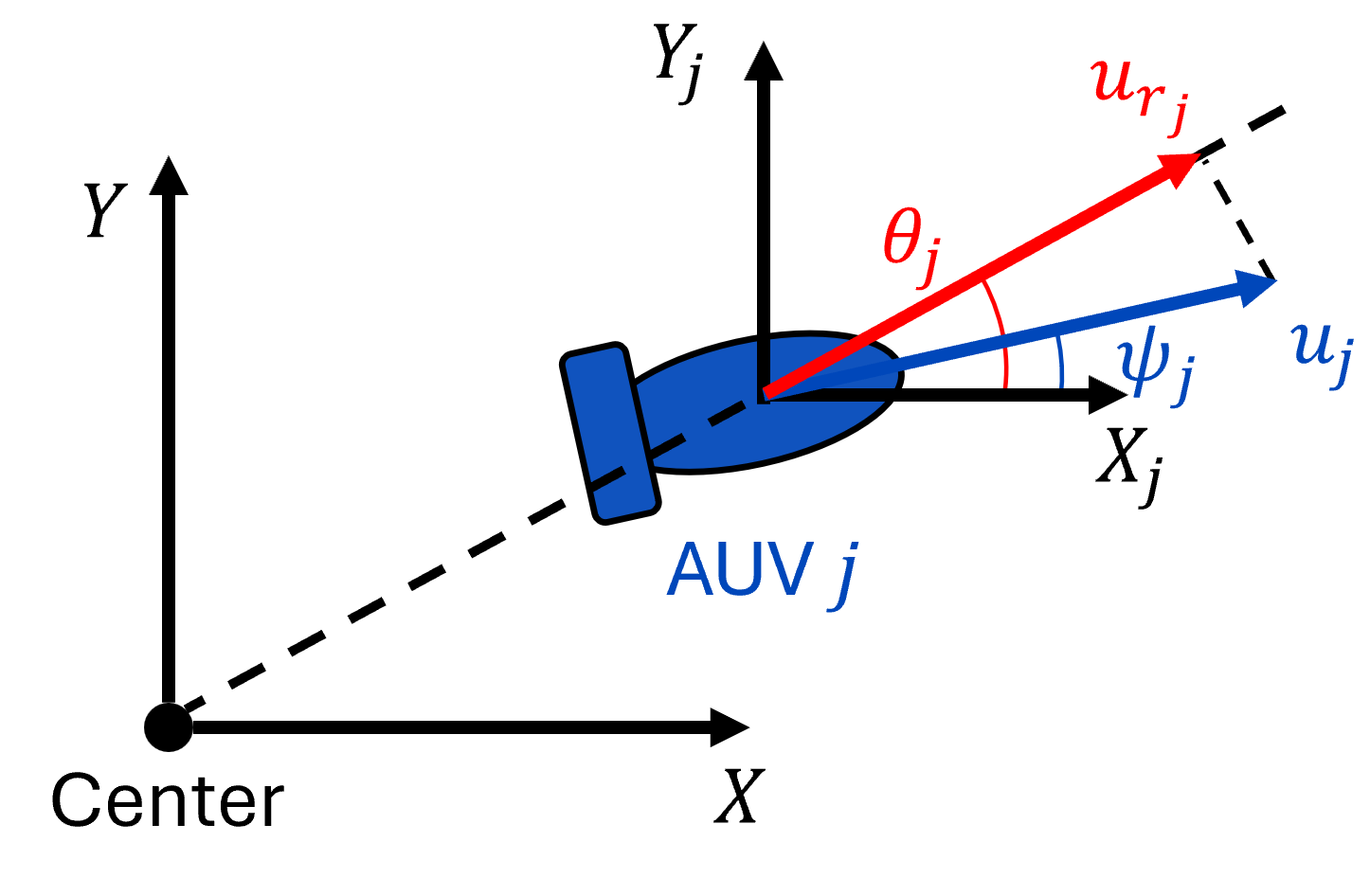}
    \caption{HEB model representation for $\theta$ estimation}
    \label{Figure_4}
\end{figure}

The HEB model is based on the estimation of the direction of the center origin, extrapolated from the history of received acoustic ranging. In particular, it uses the rate-of-change of consecutive ranges. Defining $\theta$ the direction of the axis connecting the center beacon with the AUV, the HEB control algorithms rely on the latest range measurement and the latest estimation of $\theta$. Angles and reference systems are showcased in Figure \ref{Figure_4}.

Different boundary shapes can be established for Fencing and Milling by using $\theta$ estimation. For Fencing, $\theta$ is used to make AUVs rotate towards the center of the boundary upon exiting it. During Milling, instead, $\theta$ determines the desired heading for the AUV within a specific boundary. By utilizing ranges and $\theta$ information, a large variety of closed 2-D boundaries can be created for Fencing and Milling applications. Furthermore, the latest range information is used as a sensor to activate the behavior in the case of Fencing or as a parameter to converge on the desired path for Milling behavior.

Regarding the estimation of the relative direction of the center $\theta$, as it can be inferred from Figure \ref{Figure_4}, the equation of AUV motion can be written as follows:

\begin{equation}
u_r = u \cdot cos(\theta - \psi)
\label{Equation_5}
\end{equation}

Where $u_r$ is the projected component of $u$ along the direction of the line connecting the AUV and the center of the boundary, and it is measured as the rate-of-change of two consecutive ranges.
The variable $u$ represents the velocity of the AUV, while $\psi$ is the absolute heading of the AUV. 
The following assumptions are made:
\begin{enumerate}[\indent 1.]
\item The surge speed $u$ of AUV is constant. Sway velocity is assumed null at any moment.
\item AUV is far away from the boundary center, as such for a small time window $u \Delta t \ll r$. Consequently, variation of $\theta$ is negligible within $\Delta t$.
\end{enumerate}
Equation \ref{Equation_4} can be rewritten as:

\begin{equation}
\label{Equation_6}
u_r=A\cdot\cos{\left(\psi\right)\cos{\left(\theta\right)}}+A\cdot\sin{\left(\psi\right)}\sin{\left(\theta\right)}
\end{equation}
Applying a change of variables, the equation can be expressed as: 

\begin{equation}
u_r = K_1 x_1 + K_2 x_2
\label{Equation_7}
\end{equation}
Where:

\begin{equation}
\label{Equation_8}
\begin{matrix}
x_1=&\cos{\left(\psi\right)}\\
x_2=&\sin{\left(\psi\right)}\\
K_1=&A\cdot\cos{\left(\theta\right)}\\
K_2=&A\cdot\sin{\left(\theta\right)}
\end{matrix}
\end{equation}

Given a sample of $u_r$ measurements with associated $\psi$, captured in a short interval $\Delta t$, linear regression is employed to estimate the value of $\theta$. In particular, given the values of $K_1, \ K_2$ that better fit the sample points to Eq. \ref{Equation_6}, value of $\theta$ is found as:

\begin{equation}
\label{Equation_9}
\theta=\tan^{-1}{\left(\frac{K_2}{K_1}\right)}
\end{equation}

For both Fencing and Milling, the range of the AUV with respect to the boundary center $r(t)$ is compared with $R_d(\theta)$, which represents the desired shape of the boundary or path. The simplest case is a circular boundary, where $R_d(\theta)$ is constant and equal to the radius of the circle. For other arbitrary boundaries, $R_d(\theta)$ is the polar representation of the desired path or boundary, with the origin set in the central beacon. 

For HEB Fencing, comparing $r_t$ and $R_d(\theta)$ determines whether the AUV is outside the boundary. If the AUV is outside the boundary, an absolute angle command of $\theta + 180^{\circ}$ is given. This causes the AUV to rotate and move directly towards the boundary center efficiently. When the AUV is inside the boundary, any desired behavior can be implemented. This process is detailed in Algorithm \ref{Algorithm_3}. Note that for our application, we implemented a First In, First Out (FIFO) method to collect the latest measurements since the measurements were acquired at fairly regular intervals. More in general, the collected data should be preserved in a fixed-size temporal window $\Delta t$. 

\begin{algorithm}
\caption{\enskip HEB Fencing Algorithm}\label{Algorithm_3}
\begin{algorithmic}
\State Event: Receiving new $r(t_i)$, $u_r$ and $\psi$ values
\State \textbf{Add} $u_r$ and $\psi$ to \textit{list of measurements}
\State \textbf{Remove} oldest $u_r$ and $\psi$ in \textit{list of measurements}
\State \textbf{Estimate} $\theta$ from \textit{list of measurements}
\State Calculate value of $R_{d}(\theta)$
\If {$r(t_i) > R_{d}(\theta)$}
\State $\psi_{cmd} \gets \theta + 180^{\circ}$
\Else 
\State Execute any desired in-boundary commands
\EndIf
\end{algorithmic}
\end{algorithm}

Regarding the HEB Milling procedure, the initial step involves obtaining a list of $\psi$ and $u_r$ values to estimate the value of $\theta$. Identically to the Fencing case, $R_d\left(\theta\right)$ is obtained using the estimated $\theta$ value. 
An additional parameter $\psi_d\left(\theta\right)$ is required for the Milling application; $\psi_d\left(\theta\right)$ is the absolute desired heading of an AUV following the path at a specific value of $\theta$. $R_d\left(\theta\right)$ and $\psi_d\left(\theta\right)$ are correlated and can be estimated for each desired path. For a circular path and clockwise rotation, the function of $\psi_d\left(\theta\right)$ is:

\begin{equation}
\label{Equation_10}
\psi_d\left(\theta\right)=\theta+90^{\circ}
\end{equation}

While for a square centered path, $\psi_d(\theta)$ is a step function assuming sequentially the values  $0^{\circ}$, $90^{\circ}$, $-90^{\circ}$, or $180^{\circ}$, as function $\theta$ ranges. Table \ref{Table_1} summarizes the functions of $R_d\left(\theta\right)$ and $\psi_d\left(\theta\right)$ for all clockwise paths and boundaries presented in this work. Where $m_s$ is the slope of a segment of the Isotoxal star, and $L$ is the total length of the square or the Isotoxal star.

\renewcommand{\arraystretch}{1.5}
\begin{table}[!ht]%
\begin{threeparttable}
\centering
\caption{Shape functions for circle, square, and star boundaries for clockwise Milling \label{Table_1}. The polygon functions are defined by segments.}
    \centering
\begin{tabular}{>{\centering\arraybackslash}m{0.1\columnwidth} >{\centering\arraybackslash}m{0.3\columnwidth} >{\centering\arraybackslash}m{0.45\columnwidth}}
    \hline
        \textbf{Shape} & \textbf{$\psi_d(\theta)$} & \textbf{$R_d(\theta)$} \\ \hline
        Circle & $\theta + 90^{\circ}$ & $R_0$ \\ \hline
        Square & $[0^{\circ}, 90^{\circ}, 180^{\circ}, 270^{\circ}]$ & $\begin{array}{c} \frac{L}{2\cos(\theta+\alpha)} \\ \alpha = \psi_d(\theta)+90^{\circ}\end{array} $ \\ \hline
        Isotoxal Star & $\begin{array}{c} \pm \tan^{-1}(m_s) \\ 
        \pm \tan^{-1}\left(\frac{1}{m_s}\right) \\ 
        \end{array}$ & $\begin{array}{c} \pm \left( \frac{L}{2 m_s \cos(\theta)} \right) \left( \frac{1}{\tan(\theta) \pm \frac{1}{m_s}} \right)
        \\ \pm \left( \frac{L}{2 \cos(\theta)} \right) \left( \frac{1}{\tan(\theta) \pm m_s} \right) \end{array}$ \\
    \hline
    \end{tabular}

    \begin{tablenotes}
      \small
      \item Square and Star are defined by 4 and 8 segments. Star $\psi_d$ can have two possible slopes; for each of the 2 cases, $R_d$ is described by 4 possible functions, combinations of the sign of $\pm$ as shown in Table. 
    \end{tablenotes}
  \end{threeparttable}

\end{table}

The absolute heading command to the AUV is derived as:
\begin{equation}
\label{Equation_11}
    \psi_{cmd}  = \psi_{d}(\theta) + D \  k \ (r(t_i) - R_{d}(\theta))
\end{equation}

Where $k$ is a proportional gain that determines the convergence rate of the AUV to the ideal path line. The sign of $D$ represents the rotation direction. This method enables seamless Milling motion in any direction and for any boundary shape, provided $R_d(\theta)$ and $\psi_d(\theta)$ can be uniquely defined. The entire process is showcased in Algorithm \ref{Algorithm_4}.

\begin{algorithm}
\caption{\enskip HEB Milling Algorithm}\label{Algorithm_4}
\begin{algorithmic}
\State Event: Receiving new $r(t_i)$, $u_r$ and $\psi$ values
\State \textbf{Add} $u_r$ and $\psi$ to \textit{list of measurements}
\State \textbf{Remove} oldest $u_r$ and $\psi$ in \textit{list of measurements}
\State \textbf{Estimate} $\theta$ from \textit{list of measurements}
\State Calculate $R_{d}(\theta)$ and $\psi_{d}(\theta)$ 
\State $\psi_{cmd} \gets \psi_{d}(\theta) + D \ k \ (r(t_i) - R_{d}(\theta))$
\end{algorithmic}
\end{algorithm}

The HEB model offers the flexibility to create various boundary shapes, with its main feature being the utilization of the velocity along the radial direction from the boundary center to AUV. The rate-of-change is obtainable from consecutive ranges, but it can also be directly obtained by measurement of the Doppler effect of acoustic communication (which is often already implemented in commercial acoustic transceivers). Effectively, the method presented above is equally applicable to range rates obtained from either approach. However, the use of direct Doppler-based measurement brings several significant advantages: firstly, there is no measurement delay, as the velocity is measured at the instance in which the message is received, rather than the average between two range measurements. Furthermore, for each beacon communication, each AUV can calculate its own range rate, whereas ranging measurements based on the Two Way Time Travel requires a one-to-one communication between the central beacon and each AUV. Thus, while the estimation of $r$ for each AUV can be cyclically interrogated, the range rate can be effectively collected at each beacon communication, which is of great benefit when scaling the number of AUVs involved. The drawback of direct measurement is related to the typically low measurement precision of Doppler-based estimation, which is particularly significant for small values of $u$.

\subsection{Simulations}
The algorithms were tested within a simulator designed to emulate the constraints and functionalities of the experimental platform. The AUVs' states are represented by positional coordinates $(x, y)$, velocities $(u, v)$, and heading angle $\psi$. It is not necessary to take into account the depth control in the simulator since it is decoupled from the planar control. The AUVs' motion is subject to a command that consists of a forward force $f_x$ and a heading angle $\psi_{cmd}$. An AUV’s position and orientation are updated incrementally based on the specified constant time step $\Delta t$, emulating the actual AUV dynamics. The heading control mechanism ensures a gradual adjustment towards an absolute command heading at a constant velocity $\dot{\psi}$. The below equation shows how the heading angle of an AUV is updated.

\begin{equation}
\label{Equation_12}
\psi_j(t_i)=\psi_j(t_{i-1}) + \dot{\psi} \cdot \Delta t
\end{equation}

Where $\psi_j(t_i)$ is the absolute heading of the $j-$th AUV. To calculate the AUV translation in $(x, y)$, the forces acting on the AUV are calculated. Defining $F_u$ and $F_v$ the component of the linear force acting on an AUV, with respect to surge and sway direction (body axis), those are calculated as:

\begin{equation}
\label{Equation_13}
    \begin{alignedat}{2}
        F_{uj} &= f_{xj}  && - X_{uu} \cdot u_j(t_{i-1}) \lvert u_j(t_{i-1}) \rvert  - X_u \cdot u_j(t_{i-1}) \\
        F_{vj} &= && - Y_{vv} \cdot \ v_j(t_{i-1}) \lvert v_j(t_{i-1}) \rvert - Y_v \cdot v_j(t_{i-1})
    \end{alignedat}
\end{equation}

The equation accounts for hydrodynamic drag (linear and quadratic) as well as propulsion force $f_x$. The hydrodynamic coefficients were estimated to match the ones of the real platform and are reported in Table \ref{Table_2}.

\begin{table}[!ht]%
\centering
\caption{Estimation of Hydrodynamic coefficients for AUVs\label{Table_2}}%
\begin{tabular}{>{\centering\arraybackslash}m{0.45\columnwidth}>{\centering\arraybackslash}m{0.45\columnwidth}}
\hline
$X_{u}$ & 0.1 \\
$X_{uu}$ & 4.04 \\
$Y_{v}$ & 0.1 \\
$Y_{vv}$ & 20.0 \\
\hline
\end{tabular}
\end{table}

$X_u$ and $X_{uu}$ are hydrodynamic coefficients influencing the linear and quadratic drag forces in the surge direction, while $Y_v$ and $Y_{vv}$ are hydrodynamic coefficients for the sway direction. $F_u$ and $F_v$ can be used in order to find the velocity variation in the $x$ and $y$ directions using the following equations:

\begin{equation}
\label{Equation_14}
    \begin{split}
        \Delta u_j &= \frac{F_{uj}}{m} \Delta t \\
        \vspace{3mm}
        \Delta v_j &= \frac{F_{vj}}{m} \Delta t
    \end{split}
\end{equation}

Where $m$ is the total mass (dry plus added mass) of the AUV. Eventually, after updating the velocity, the position of $x(t)$ and $y(t)$ can be calculated for each AUV as follows:

\begin{equation}
\label{Equation_15}
\begin{split}
x_j(t) &= x_j(t_{i-1}) + (u_j(t) \cdot cos(\psi_j) + v_j(t) \cdot sin(\psi_j)) \Delta t\\
y_j(t) &= y_j(t_{i-1}) + (u_j(t) \cdot sin(\psi_j) - \ v_j(t) \cdot cos(\psi_j)) \Delta t
\end{split}
\end{equation}

Regarding acoustic communication, delays, and time quantization are introduced in order to emulate the acoustic channel effects in the simulator. A delay and quantization of one second are incorporated into the estimation of ranging information for each AUV within the simulation. This closely mirrors the time required for each packet transmission in real-world scenarios.

\subsubsection{Fencing Simulations}
\label{subsec:Fencing}
Fencing simulations were conducted for scenarios involving one, two, and three AUVs for 1000 $s$. The primarily tested algorithm was HEB Fencing. The parameters selected are shown in Table \ref{Table_3}. The boundaries tested are circle, square and Isotoxal star. Circle boundary has a radius of 30 $m$, the square boundary has sides of length 60 $m$, while the Isotoxal star 4 vertexes are located at $\pm 30$ $m$, and its concave points are located at $(\pm10, \pm10)m$, depending on the quadrant. The Isotoxal star boundary is presented to test the algorithms with a complex and concave shape.

\begin{table}[!ht]%
\centering
\caption{Parameters used for HEB Fencing simulations\label{Table_3}}%
\begin{tabular}{>{\centering\arraybackslash}m{0.45\columnwidth}>{\centering\arraybackslash}m{0.45\columnwidth}}
\hline
$f_x$ & 0.5 $N$ \\
Quantization \& Range Delay & 1 $s$ for each AUV in simulation\\
Length of List & 5 \\
\hline
\end{tabular}
\end{table}

In order to evaluate the performance of Fencing, the metrics used are \textit{maximum range error} (MRE), \textit{mean peak error} (MPE), and \textit{average return time} (ART). The position of the AUV can be expressed in the polar coordinates $(r(t), \theta(t))$, with respect to the central beacon. Let us define "dip" as a section of trajectory that lasts from the instance in which the AUV exits the boundary to the instance in which it returns within it. For experimental tests, occasionally, the AUV approaches the boundary but is not able to cross it. In such cases, we use the instance in which we get the minimal range from the center, still outside the boundary, as the delimiting point of the dip. 
Each dip is consequently limited in a $[t_i, t_{i+1}]$ time range, with $i \in [1,N]$. The overshoot peak $p_i$ of the $i$-th dip is:

\begin{equation}
    p_i = \max_{[t_i, t_\text{i+1}]} \left[ r(t) - R_d(\theta(t)) \right]
    \label{Equation_16}
\end{equation}

Where $R_d(\theta(t))$ is the desired position of the AUV at the time $t$. Consequently, we can define the \textit{maximum range error} (MRE) as: 

\begin{equation}
\label{Equation_17}
    \text{MRE} = \max_i p_i
\end{equation}

Another interesting value is the \textit{mean peak error} (MPE), defined as the average of the maximum overshoots of the AUVs recorded during the test. With this definition, MPE can be calculated calculated as:

\begin{equation}
\label{Equation_18}
    \text{MPE} = \frac{\sum_i p_i}{N}
\end{equation}

Finally, the return time $\Delta t_{ret}(i)$ is the time taken by AUVs to come back to the boundary after leaving it. The \textit{average return time} (ART) is calculated as:

\begin{equation}
\label{Equation_19}
    ART = \frac{\sum \Delta t_{ret}(i)}{N}
\end{equation}

\begin{figure}[!t]
    \centering
    \includegraphics[width=0.5\textwidth]{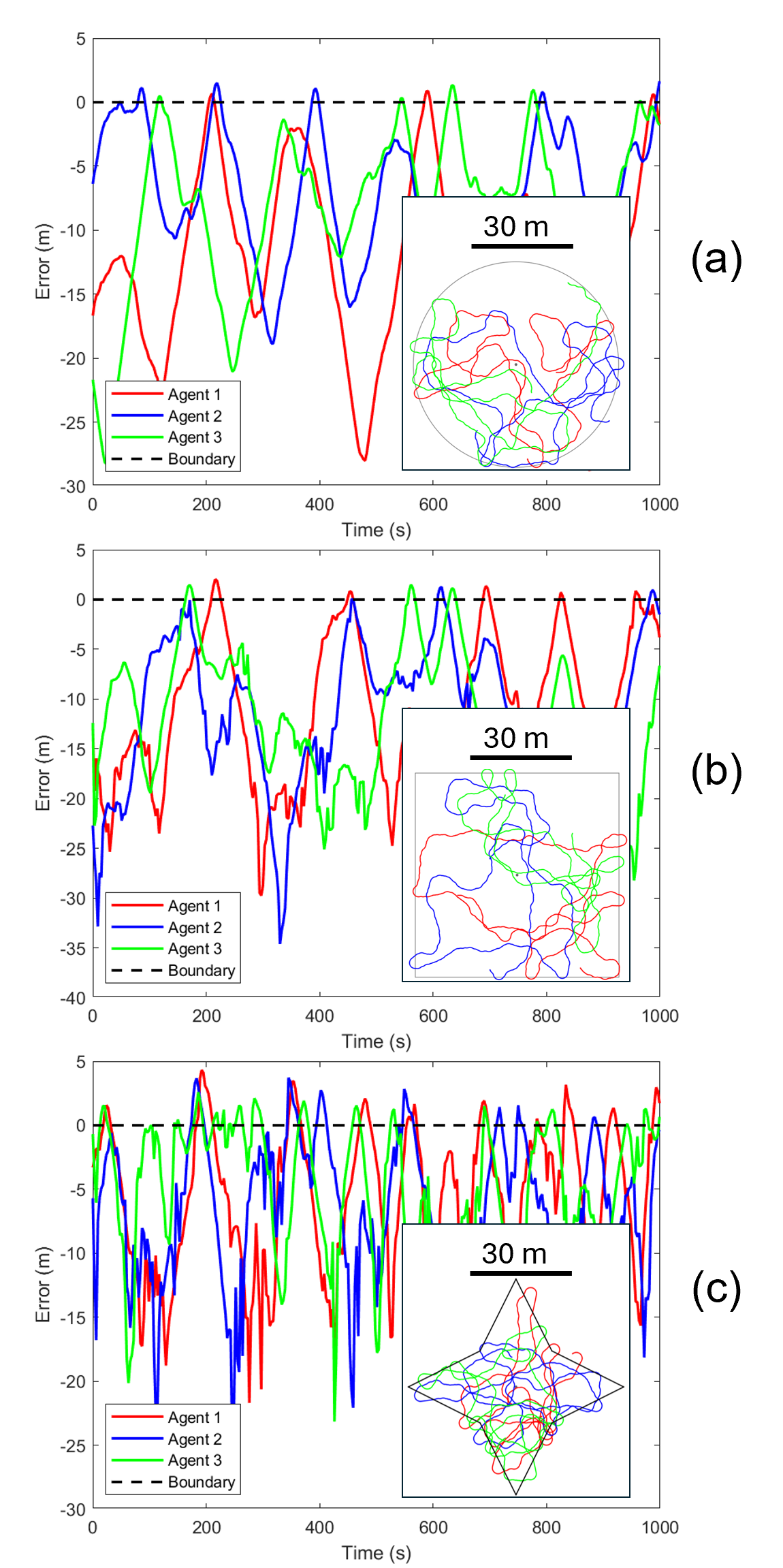}
    \caption{HEB Fencing on 3 boundaries for 3 AUVs. (a) Circle, (b) Square, (c) Isotoxal Star}
    \label{Figure_5}
\end{figure}

Figure \ref{Figure_5} shows HEB Fencing for three AUVs on the three boundaries. Figure \ref{Figure_5} (a) shows Fencing on the circle boundary. Notably, as the AUVs slightly move beyond the boundary line, they swiftly adjust their trajectories, thereby limiting the maximum distance reached from the center. Figure \ref{Figure_5} (b) showcases Fencing on the square boundary, where we can observe similar overshoots as in the circle boundary. Figure \ref{Figure_5} (c) shows HEB Fencing on Isotoxal star boundary, where, contrary to the circle and square boundaries, it has a greater overshoot. The increased error can be attributed to the sharp corners of the Isotoxal star, which demand more precise path adjustments from the AUVs. Specifically for the circle boundary, the simulations were also performed for one and two AUVs.

The MRE for single and double AUVs simulation are 1.458 $m$ and 1.461 $m$ compared with 1.905 $m$ error obtained in three AUVs simulations. Similarly, we observe an increase in MPE and ART as the number of AUVs increases. This can be attributed to the growing range delay when deploying a larger number of AUVs. Higher surge velocities also resulted in greater errors, as for an equal time interval the AUV covered larger distances before receiving a ranging update. However, it is important to note that the errors are independent of the boundary sizes. Hence the proportional error, compared to boundary size, quickly reduces when the behavior is applied to larger boundaries.

RVB Fencing was also simulated with the same parameters used for the HEB Fencing on the circle boundary. Additionally, the value of $\Delta \psi$ in RVB Fencing is set to 20$^\circ$. The algorithm is effective in constraining the AUV to the boundary but with a larger turning curvature compared to HEB Fencing. The MRE for three AUVs reaches 4 $m$ in RVB Fencing, which is twice the error experienced in HEB Fencing. Furthermore, the ART in RVB Fencing is 22.4 $s$ which also almost double the magnitude seen in HEB Fencing. This result is expected since the RVB algorithm requires several rotations to correct its heading before stabilizing around the optimal return value. Those initial corrections delay the return and consequently allow the AUV to move further away from the boundary, thus increasing both MPE and ART. The conducted simulations successfully validated the performance of HEB and RVB Fencing. Results are further summarized and discussed in Table \ref{Table_6}.

\subsubsection{Milling Simulations}
\label{subsec:Milling}
\begin{figure}[!t]
    \centering
    \includegraphics[width=0.5\textwidth]{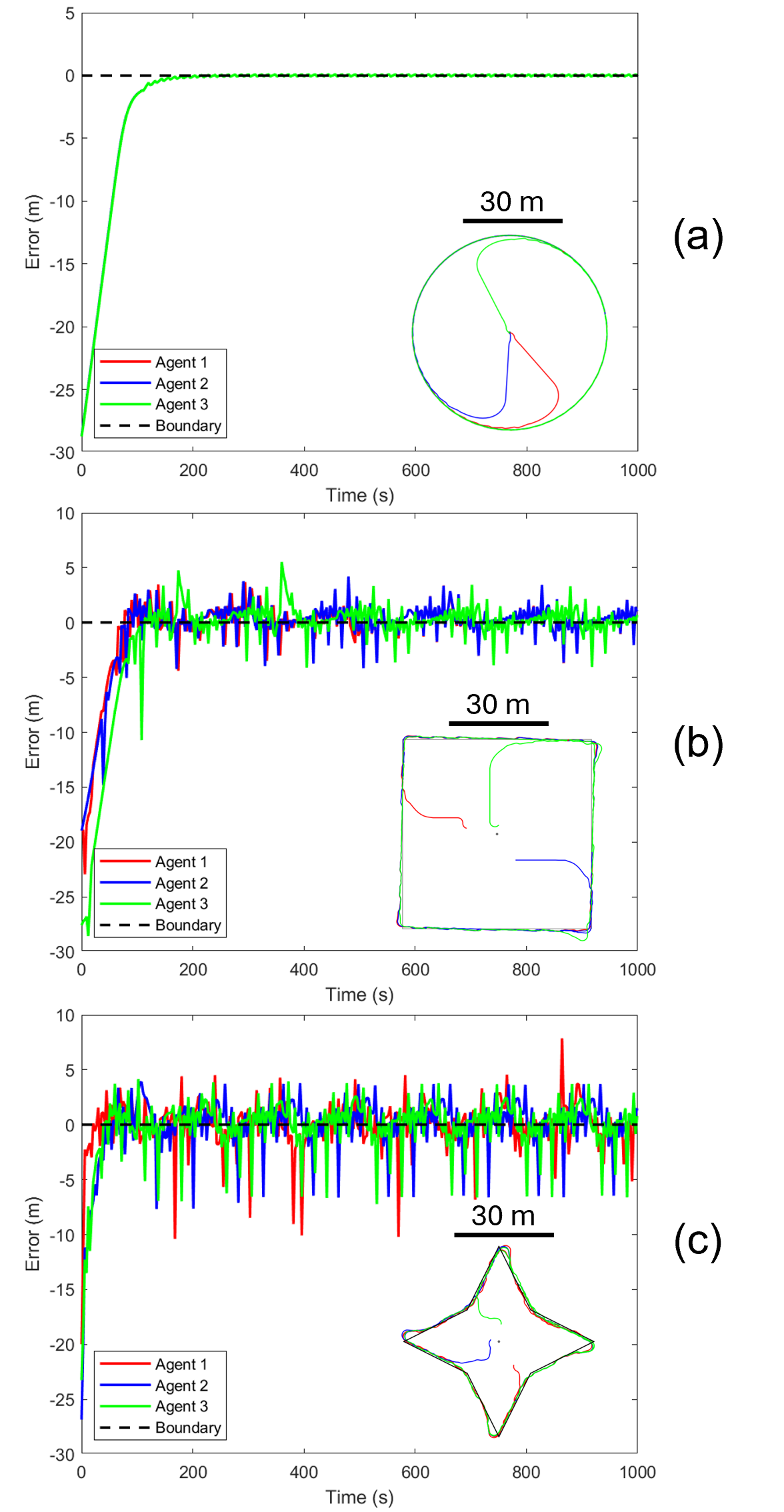}
    \caption{HEB Milling on 3 boundaries. (a) Circle, (b) Square, (c) Isotoxal Star}
    \label{Figure_6}
\end{figure}

HEB Milling simulations are conducted for 1000 $s$ on the three paths with the same dimensions and parameters as in the Fencing Section \ref{subsec:Fencing}. For the simulation, we selected a convergence parameter $k=20$ and $k_{rate}=0$. The results are depicted in Figure \ref{Figure_6}.

In order to evaluate the performance of Milling, the metrics used are the \textit{maximum range error} (MRE), the \textit{accuracy} $\overline{\mu}$ and the \textit{precision} $\overline{\sigma}$. Each desired path can be defined by using polar coordinates. Taking the angular position $\theta$ as an input variable, the radius of the desired path is $R_d(\theta)$. At each instance of the test, an error $e(t)$ can be computed as:

\begin{equation}
\label{Equation_20}
    e(t) = r(t) - R_d(\theta(t))
\end{equation}
Where $r(t)$ is the real radial position of the AUV. The MRE for Milling can be obtained as a result of Eq. \ref{Equation_16} over the whole testing time.

Another interesting error value is the mean error $e_m(t)$, computed over the mean radius $\overline{r}$ achieved during the test:

\begin{equation}
\label{Equation_21}
    e_m(t) = r(t) - \overline{r} , \ \quad \overline{r} = \frac{\sum r(t)}{N}
\end{equation}

Where $N$ is the number of total instances of radial position stored in the experiment sample.

For the circle boundary experiments, these values assume a simple solution since $R_d(\theta) = R_0, \ \forall \theta$. Finally, the bias on the ranging, defined as the \textit{accuracy} $\overline{\mu}$, and the standard deviation of the mean error, defined as the \textit{precision} $\overline{\sigma}$, are obtained as:

\begin{equation}
\label{Equation_22}
    \begin{split}
        \overline{\mu}(e(t)) &= \frac{\sum \left[r(t) - R_d(\theta(t))\right]}{N} = \overline{r} - R_0 \\
        \overline{\sigma}(e_m(t)) &= \sqrt{\frac{\sum {e_m}^2(t)}{N}}
    \end{split}
\end{equation}

Figure \ref{Figure_6} (a) showcases that HEB Milling on a circle boundary is stable and consistent where the MRE reached 0.088 $m$. It is observed that there is no overshoot in range, and AUVs' trajectories require, on average, around 170 $s$ to reach and settle on the path line. In Figure \ref{Figure_6}(b), the AUVs occasionally overshoot at the corners of the square boundary due to the $90^{\circ}$ rotations. The AUVs' trajectories follow up closely the desired Milling path, with minimal error for the majority of its extent. The maximum divergence occurs at the corners, where the MRE reaches up to 5 $m$. The overshoot is greater in the Isotoxal star boundary seen in Figure \ref{Figure_6} (c) due to sharper turns at vertexes when compared with the square boundary. The MRE reached by AUVs is about 8 $m$. While corner discontinuity introduces overshoots in the AUVs trajectory, it can be qualitatively observed that the algorithm is able to quickly compensate and return the AUV on the path, with varying settling time depending on the angle of the corner and on the speed of the AUV.

\begin{figure}[!t]
    \centering
    \includegraphics[width=0.5\textwidth]{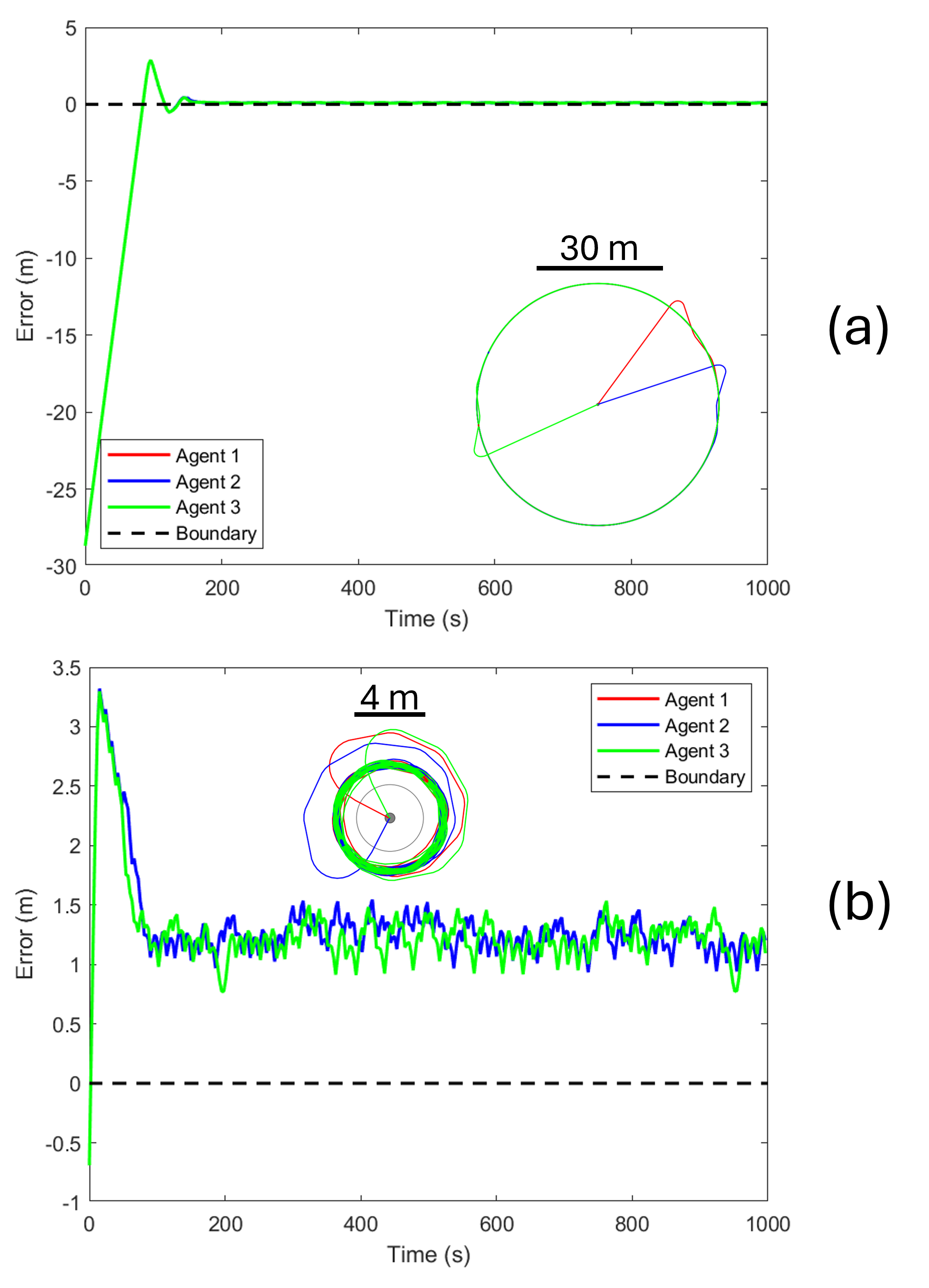}
    \caption{RVB Milling for three AUVs on the circular boundary. (a) range error and positions of AUVs on 30 $m$ radius circular boundary, (b) 2 $m$ radius circle}
    \label{Figure_7}
\end{figure}

As explained in Section \ref{subsec:RVB}, RVB Milling can only generate a circular path. The parameters used in the simulation are $k=20$ and $k_{rate}=0$. We tested for $R_0=$ 30$m$ and 2$m$.
Similar to the HEB, RVB Milling was tested for one, two, and three AUVs. Figure \ref{Figure_7} shows the resulting paths and quantitative results are showcased in Table \ref{Table_7}. The AUVs exhibit a similar response to the HEB Milling on a circle path. However, we can observe a larger MRE, up to 3 $m$. As expected of the algorithm, due to condition (d) in Section \ref{subsec:RVB}, the rotation initiates only after the AUV exits the area delimited by the path, thus causing the overshoot. The settling time in the RVB Milling is about 160 $s$, which is similar to HEB Milling, even with the presence of the aforementioned initial overshoot in RVB Milling. For smaller radii (2 $m$), we observe that, in the RVB Milling, the mean radius $\overline{r}$ stabilizes at 1.221 $m$ beyond the boundary line, thus never fully reaching the desired state. In conclusion, simulated results confirm the efficacy of the RVB Milling.

\section{H-SURF AUV}\label{H-SURF AUV}
H-SURF system consists of three different types of robots: Floater, Sinker, and Fish (AUV) \cite{Iacoponi2022}. Specifically, the H-SURF system is composed of a single Floater-Sinker and several Fish-AUVs, which will be referred to as Agents. The Floater is an Unmanned Surface Vehicle (USV), and acts as a gateway between the operators, to which it communicates through WiFi, and the underwater system, reached through acoustic communication. The Sinker serves as a passive component in the system, as it lacks any propulsion mechanism and is tethered to the Floater. 

\begin{figure}[!hbt]
    \centering
    \includegraphics[width=0.4\textwidth]{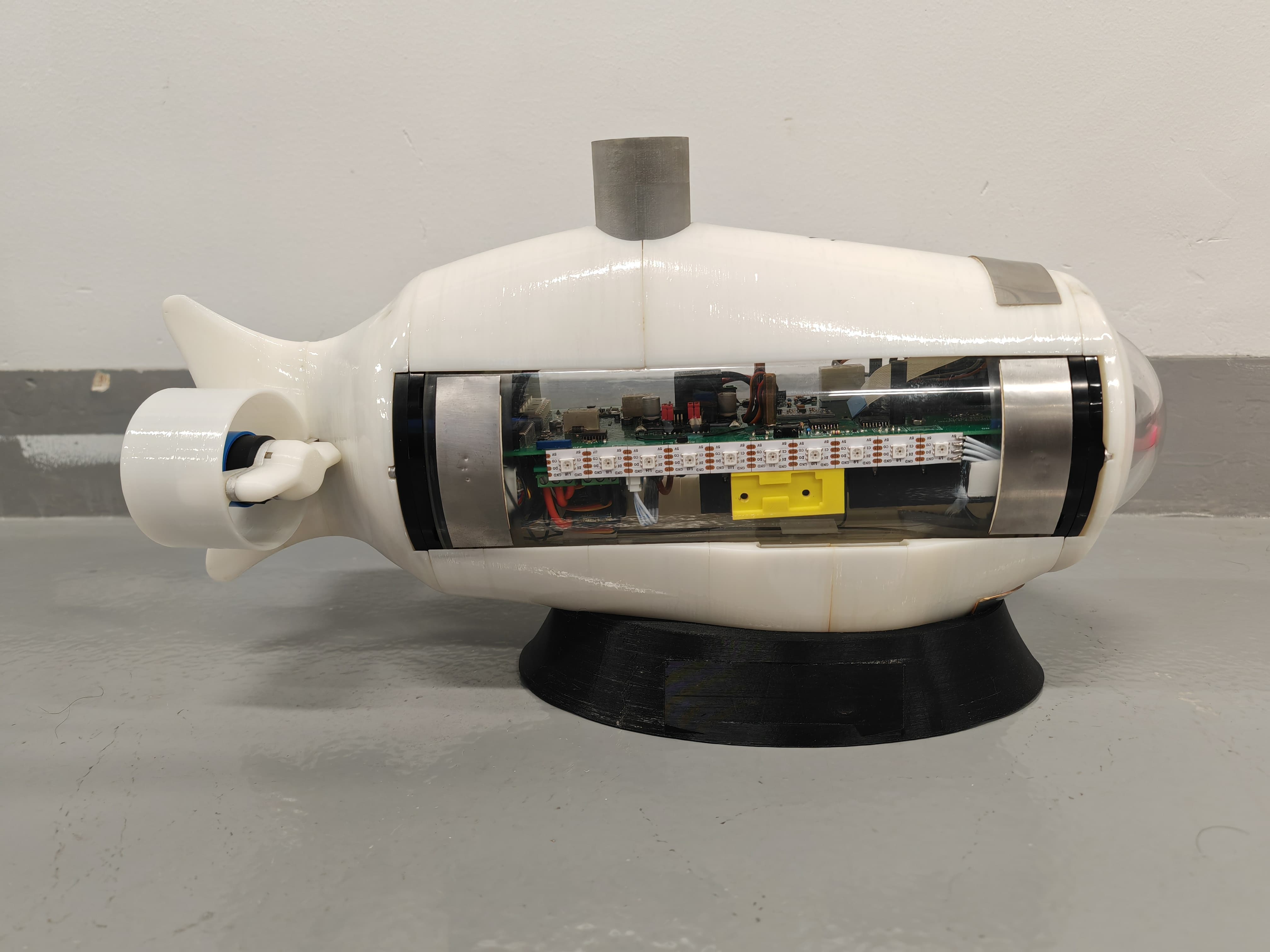}
    \caption{H-SURF AUV, a single Agent. Note the Succorfish V3 Acoustic modem mounted on top and the 2 rear propellers on the back providing surge and heading correction.}
    \label{Figure_8}
\end{figure}

Figure \ref{Figure_8} shows the Agent, while Table \ref{Table_4} provides an outline of its main components and specifications. Each Agent mounts a 9-DoF Inertial Measurement Unit (IMU) that is used to determine the absolute heading and a pressure sensor to determine depth. The Agents can exert control over 3 DoF with 3 brushless motors. Thus, it is an under-actuated system: surge and yaw are controlled by a couple of motors on the rear side of the Agent, while heave is managed by a third vertical bottom-mounted motor. The movement of Agents is controlled by an input consisting of surge force, absolute heading, and absolute depth, consistent with the command described in the model and in the simulations. Low-level control, described in detail in \cite{Iacoponi2022}, converts the input into motor signals. Furthermore, each Agent has an acoustic modem that enables communication with the other Agents and the Floater-Sinker. The modem transceiver used is a Delphis V3 which has a single-channel acoustic communication that can measure ranges using ping protocol. Essential specifications of Delphis V3 are shown in Table \ref{Table_5}. Regarding the behaviors tested in this study, communication is established solely between the Floater and the Agents. Hence, the Floater acts as a single beacon that provides necessary information for controlling the path of the Agents. 

\begin{table}[!ht]%
\centering
\caption{Specifications of H-SURF AUV\label{Table_4}}%
\begin{tabular}{>{\raggedright\arraybackslash}m{0.25\columnwidth} m{0.65\columnwidth}}
\hline
Size (cm) & 45 (L) x 16 (W) x 23 (H) \\
Dry Weight & 4 kg \\
Thrusters & Brushless Motors (Frame: ROV with a Single Vertical Thruster) \\ 
Computer & Raspberry Pi 4B+ \\
OS & Ubuntu 22.04 + ROS2 Humble \\
Micro-controller & ESP32 \\
Acoustic Modem & Succorfish Delphis V3 \\
Heading Sensor & IvenSense MPU9250 \\
Pressure/Depth Sensor  & TE Connectivity MS5837-30BA \\
Echo-Sounder & MaxBotix MB7040 I2CXL-MaxSonar-WR \\
\hline
\end{tabular}
\end{table}

\begin{table}[!ht]%
\centering
\caption{Essential specifications of acoustic modem Delphis V3\label{Table_5}}%
\begin{tabular}{>{\raggedright\arraybackslash}m{0.65\columnwidth} m{0.25\columnwidth}}
\hline
\textbf{Specification} & \textbf{Value} \\ \hline
Maximum Range (Sea Water) & 2 $km$ \\
Maximum Range (Fresh Water) & 3.5 $km$ \\
Ranging Increment & 4.7 $cm$ \\
Package Size & 64 Bytes \\
Package Transmission Time & $\sim 1 \ s$ \\
\hline
\end{tabular}
\end{table}

\begin{figure}[!hbt]
    \centering
    \includegraphics[width=0.5\textwidth]{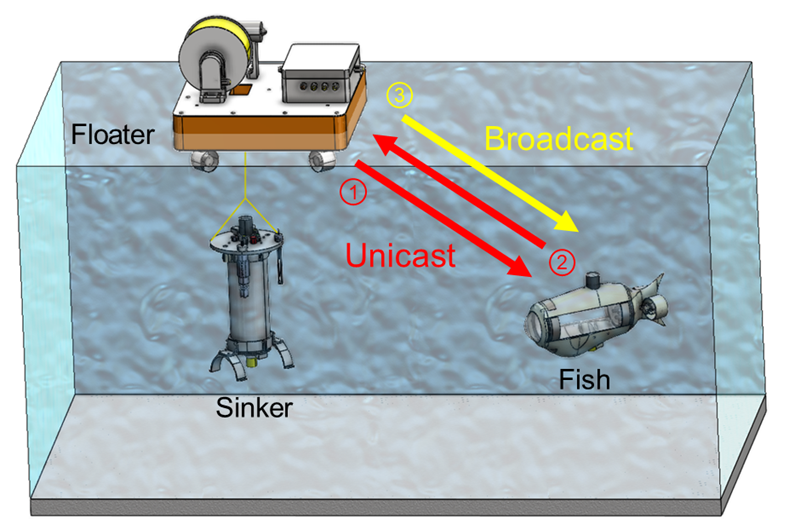}
    \caption{H-SURF setup and acoustic communication representation. For each Agent the protocol sequence consists of: one ping from the Floater, one ping response from the Agent, and one broadcast package from the Floater.}
    \label{Figure_9}
\end{figure}

Communication between Floater and Agents occurs through unicast and broadcast protocols, similar to \cite{Matsuda2022}. Unicast is a one-way transmission between Floater and Agent, while broadcast allows the Floater to transmit data to all Agents at once. The H-SURF setup, along with a representation of the communication methods, is shown in Figure \ref{Figure_9}. 

The communication is conducted in a sequential manner, alternating between unicast and broadcast transmissions. At first, the Floater sends an unicast message to the desired Agent. An acknowledgment message is then expected from the Agent to confirm the reception. If the communications succeed, the Floater obtains an estimation of the range between the platforms, assuming a speed of the acoustic waves underwater of 1500 $m/s$. The value of the range is stored, and the rate-of-change of the Agent's ranges is computed using the current and the previous range. Finally, a broadcast message containing information regarding all the active Agents is sent to allow each robot to update its own internal knowledge. We can notice that the frequency of the Agent's own range reception decreases as the number of Agents increases since the evaluation of ranges is obtained through a two-way unicast transmission.

\section{Pool Experiments}\label{Pool Experiments}
\begin{figure}[!hbt]
    \centering
    \includegraphics[width=0.4\textwidth]{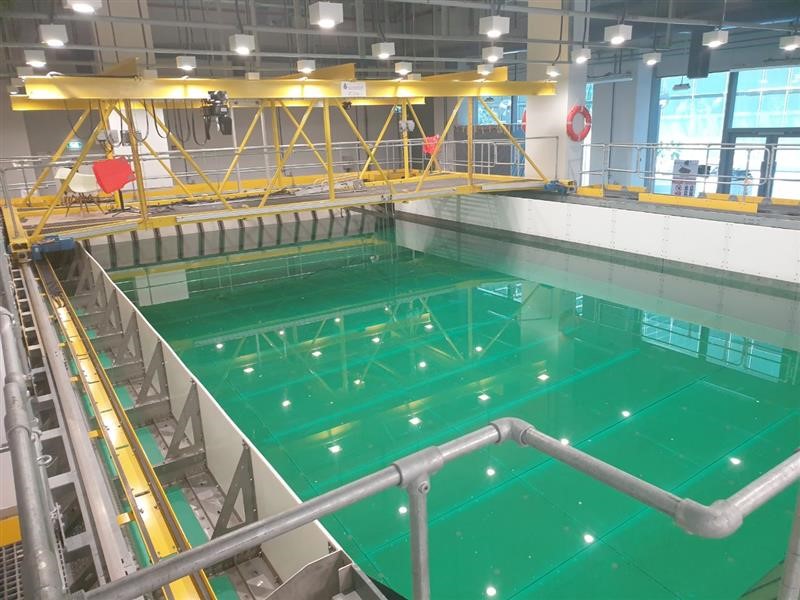}
    \caption{Pool in Khalifa University Marine Lab, size 12 x 8 x 2 $m$. On top, set of 12 tracking cameras. The long side of the pool features a metal grid through which the water is circulated by the underwater current generator.}
    \label{Figure_10}
\end{figure}

Testing in a controlled environment is conducted for the implementation of the Fencing and Milling on H-SURF robots. Experiments are performed in the Marine Lab pool at Khalifa University. The pool measures $ 8 $m$ (W) \times  12 $m$ (L) \times  2 $m$ (D)$
, and is shown in Figure \ref{Figure_10}. The pool features an underwater current generator, which allows a controlled constant laminar flow at the desired velocity alongside the width of the pool. This feature can be effectively used to emulate underwater currents.

In order to evaluate the performance of Fencing and Milling, three distinct tracking methods were utilized: 
\begin{enumerate}[\indent 1.]
    \item acoustic ranges obtained from the modems' Time of Flight
    \item tracking system above the pool (OptiTrack)
    \item monocular 2-D tracking on video (Kinovea)
\end{enumerate}
The first method uses data from acoustic ranges collected onboard the Agents. To extract the absolute position ground-truth of the Agents, a tracking system and monocular 2-D tracking were used. The tracking system utilizes 12 OptiTrack cameras placed above the pool to detect a set of reflective markers. Similarly, monocular 2-D tracking uses a single camera mounted poolside to record the test. The collected video is then processed with the Kinovea software, allowing to reconstruct the Agents' position, using the markers as a reference alongside the known dimensions of the pool. A hydrofoil-shaped fin was designed and rigidly mounted on top of the Agents in order to keep the markers above the water surface, as well as to maintain the Agents at a constant depth during the tests.

\subsection{Evaluation of Tracking Methods}
\begin{figure}[!hbt]
    \centering
    \includegraphics[width=0.4\textwidth]{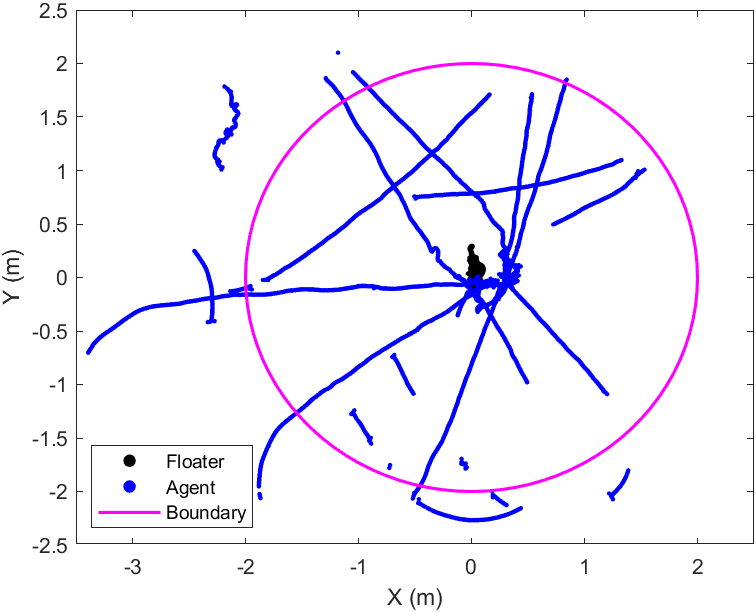}
    \caption{Position collected for Fencing with one Agent, collected by the Optitrack tracking system. Note the frequent discontinuity, especially when roaming outside the Fencing boundary.}
    \label{Figure_11}
\end{figure}

The main reason two separate ground truths have been implemented is to address the specific limitations of each method. While the Optitrack system is a well-established and accurate method to capture the pose of each Agent, its implementation in the pool severely limits the tracking volume and consistency due to the positioning limitation of the cameras in the pool lab and to the effects of water reflection on tracking markers acquisition. These limitations result in strongly discontinuous datasets. Acoustic and monocular 2-D tracking instead required validation. Validation was performed on the single Agent Fencing test. Figure \ref{Figure_11} shows the position data that has been extracted from the tracking system. From the data retrieved, we can assess that the area covered by the tracking system is not sufficient to properly show the complete movement of the Agent. Also, the data is discontinuous, mostly due to the reflection of the OptiTrack system's infrared lights on the water surface. Hence, the data obtained from the tracking system can not be used on its own to analyze the behaviors. However, thanks to the high accuracy of the gathered data, it can be used as a baseline to evaluate other tracking methods.

\begin{figure}[!hbt]
    \centering
    \includegraphics[width=0.4\textwidth]{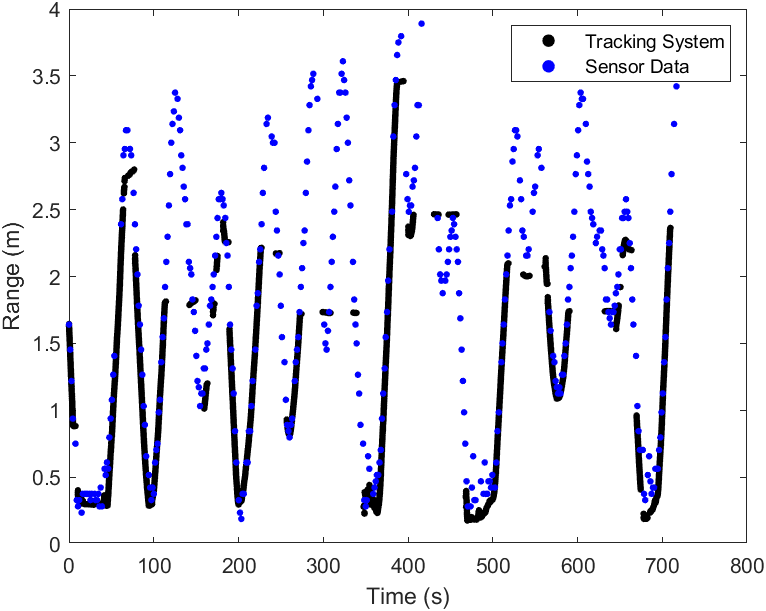}
    \caption{Comparing the Optitrack tracking system data with the acoustic sensor data collected onboard the Agent. Data are synchronized within the ROS environment. Within noise oscillation, it can be clearly noted a very close match between the two of the ranges. Ranges from Optitrack are calculated known the absolute positions of Floater and Agent.}
    \label{Figure_12}
\end{figure}

The discontinuous data from the tracking system is compared with the ranging sensor and video analysis data. It is clear, from Figure \ref{Figure_12}, that the data points from the ranging sensor closely follow the data points from the tracking system, with a root-mean-square error (RMSE) of 0.2 $m$. Furthermore, acoustic ranging is generally continuous during testing. We can conclude that ranging collected during tests can be used directly for quantitative analysis of the performance.

\begin{figure}[!hbt]
    \centering
    \includegraphics[width=0.4\textwidth]{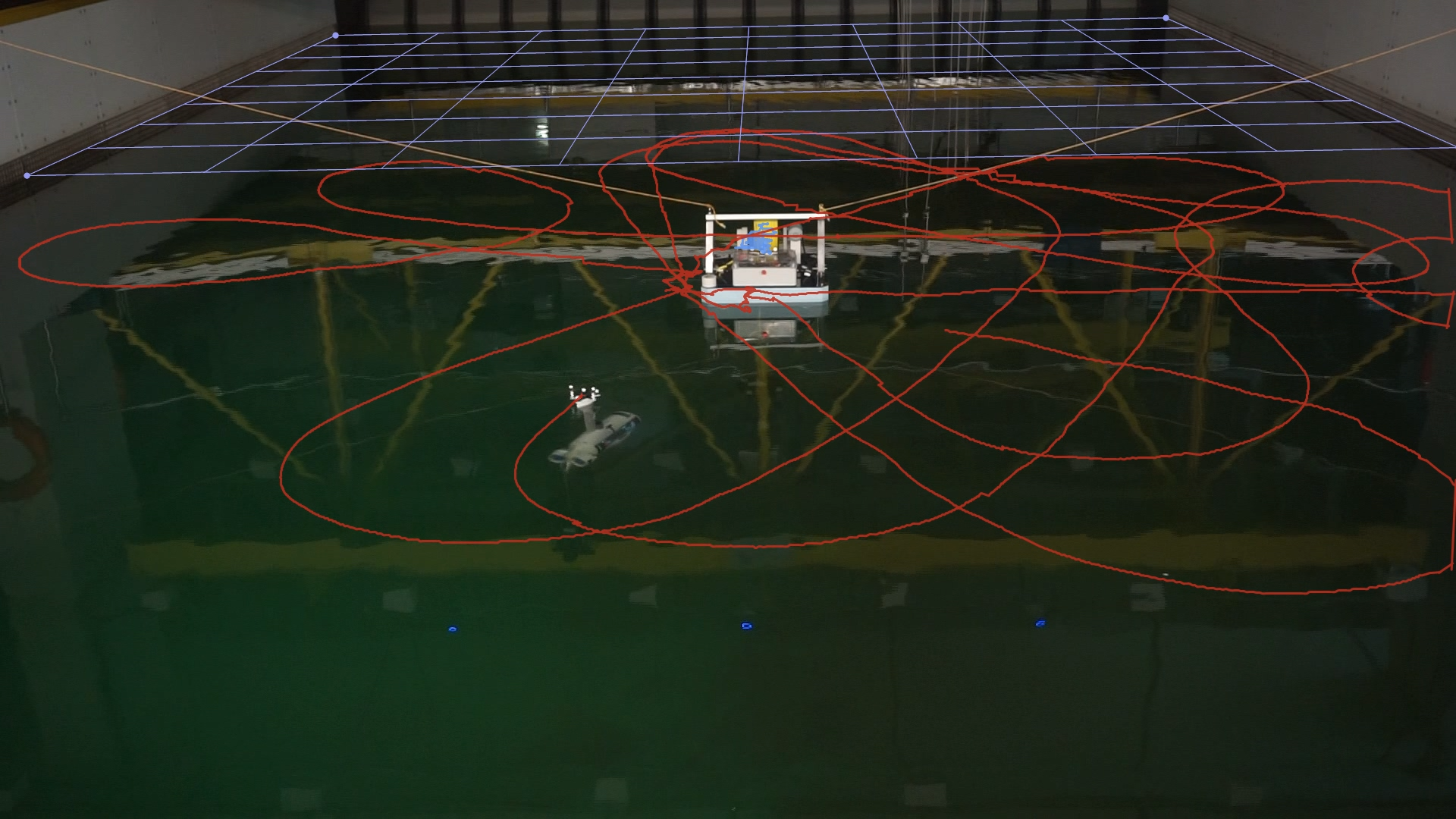}
    \caption{Video post-processing of Fencing pool experiment with one Agent. Grid created on the water surface is shown in blue, while the motion of Agent during the experiment is depicted with red lines}
    \label{Figure_13}
\end{figure}

Figure \ref{Figure_13} shows the output on video of the monocular 2-D tracking with Kinovea for the same test. There is minimal data discontinuity, corresponding to sections where the Agent exits the camera's field of view. Ranges obtained from video analysis were compared with tracking system ranges, similar to the comparison made with sensor data against the tracking system. We noted a higher deviation between data, especially when the Agent is far away from the camera. The RMSE of the video analysis method is 0.37 $m$. Monocular 2-D tracking holds the advantage of fully reconstructing the trajectory and thus is presented primarily for qualitative analysis.

\subsection{Fencing Pool Tests}
\label{subsec:FenPool}

\subsubsection{Static Water Fencing}
\begin{figure*}[!hbt]
    \centering
    \includegraphics[width=1\textwidth]{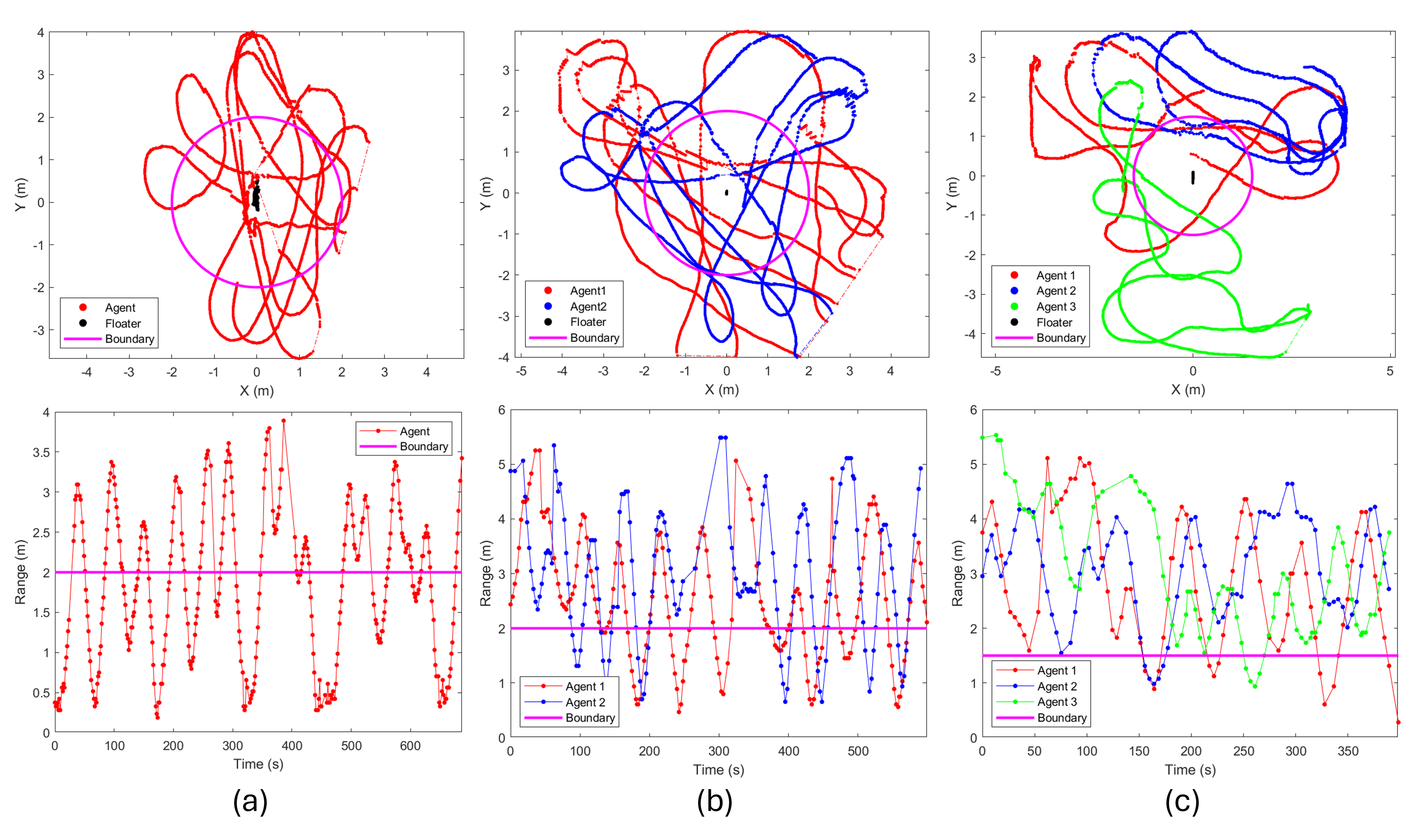}
    \caption{Fencing with no underwater current. Above, the reconstructed path, with the boundary in purple and the effective position of the floater in black. Below, the relative acoustic ranges of Agents from Floater. a) Single Agent, b) Two Agents, c) Three Agents and reduced $f_x=0.3N$}
    \label{Figure_14}
\end{figure*}

HEB Fencing experiment was implemented involving one and two Agents, using a circular boundary of 2 $m$ in radius, and three Agents with a 1.5 $m$ radius. 
For the tests with one and two Agents, $f_x$ was set at 0.5$N$, while for three Agents it was reduced at 0.3$N$. A difference from the simulated test is that, while within the boundary, the Agents are commanded to maintain the last heading. This allows us to easily identify the effects of the algorithm since any rotation is either a disturbance or a result of the Fencing protocol.
Figure \ref{Figure_14} showcases the paths followed by the Agents in the various tests and associated ranges, as recorded by the individual Agents. 
It is clear that the Agents outside the boundary are performing effective rotations to come back toward the boundary. Especially in the single-Agent case, it is clearly visible that, upon exiting the boundary, the Agent rotates in order to approximately point the center. The same can be observed for two Agents, albeit with larger overshoots. In three Agents, results are not as clear, but we can still observe that all Agents are eventually confined and that the protocol is effective in limiting the Agents operational area. We applied the metrics defined in Section \ref{subsec:Fencing}: quantitative results are showcased in Table \ref{Table_6}.

\subsubsection{Fencing with Underwater Current}
\begin{figure}[!hbt]
    \centering
    \includegraphics[width=0.4\textwidth]{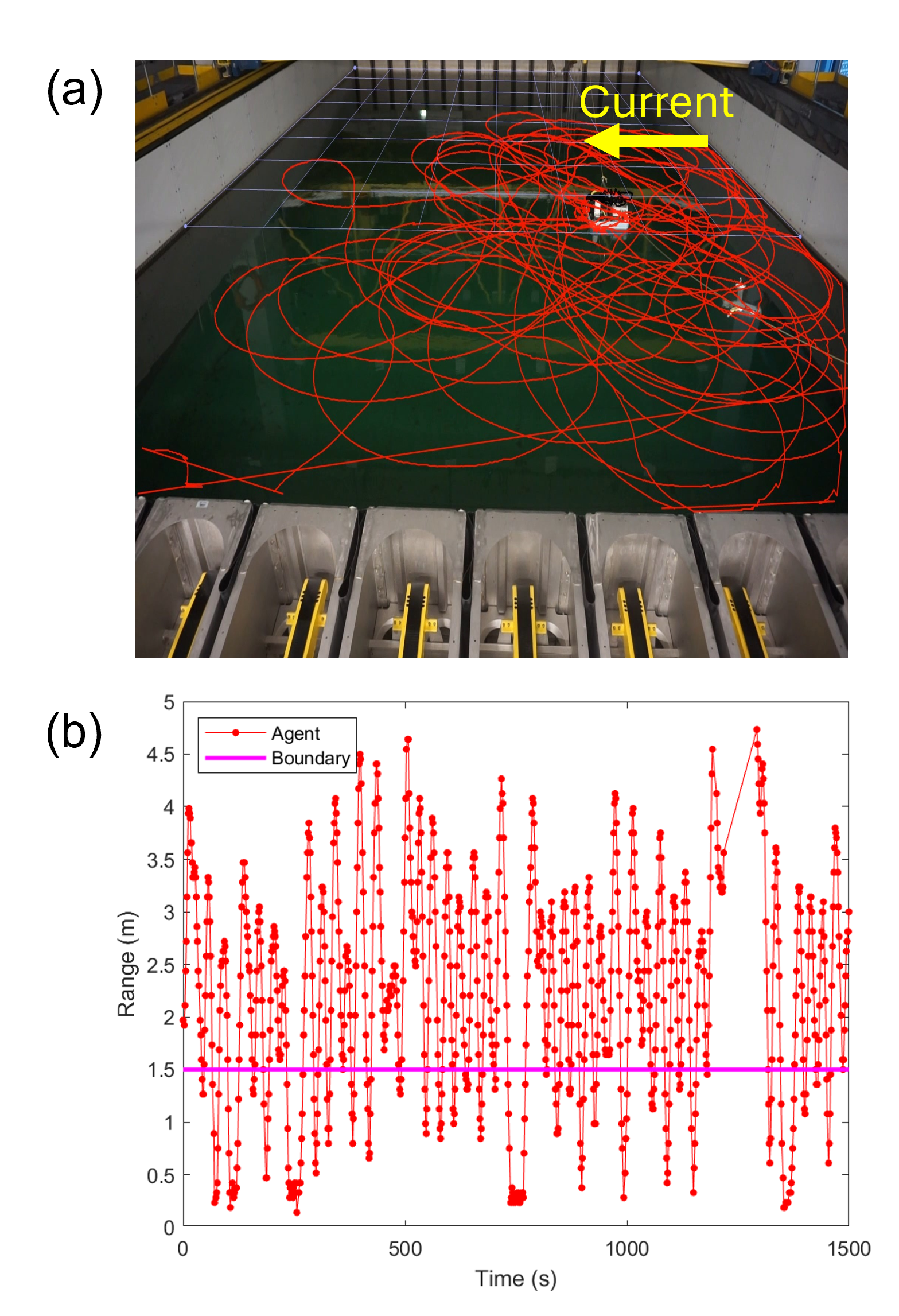}
    \caption{Fencing of one Agent, with a constant stream of underwater current at 0.08$m/s$, full test represented. a) Path tracing based on a monocular camera. b) Relative ranges collected by the Agent}
    \label{Figure_15}
\end{figure}

In order to emulate a realistic scenario, such as the presence of an underwater current or moving beacon, Fencing behavior was tested with the addition of underwater currents inside the pool. The Floater was held in position with two ropes, tightly connected to the pool walls, at approximately 2 $m$ from the outlet wall. A single Agent Fencing was initiated, and the underwater current generator was activated with a set stream velocity of 0.08 $m/s$, not dissimilar to oceanic underwater currents as seen in \cite{Statnikov}. To ensure that the Agent can outrun the underwater current while rotating, the $f_x$ of the Agent is set at 1$N$, and a circle boundary with a radius of 1.5 $m$ is set.
The results of the experiment, along with the direction of underwater current, are shown in figure \ref{Figure_15}.  Figure \ref{Figure_15} (a) clearly shows that overshoots outside the boundary are more prevalent on the left side due to the underwater current effect. Ranging measurements (Figure \ref{Figure_15} (b)) showcase an MRE of 3 $m$. The successful containment of the Agent by the Fencing behavior, with active underwater current, implies that the utilization of Fencing in open ocean water is applicable, even in the presence of moderate underwater currents.

\begin{figure}[!hbt]
    \centering
    \includegraphics[width=0.4\textwidth]{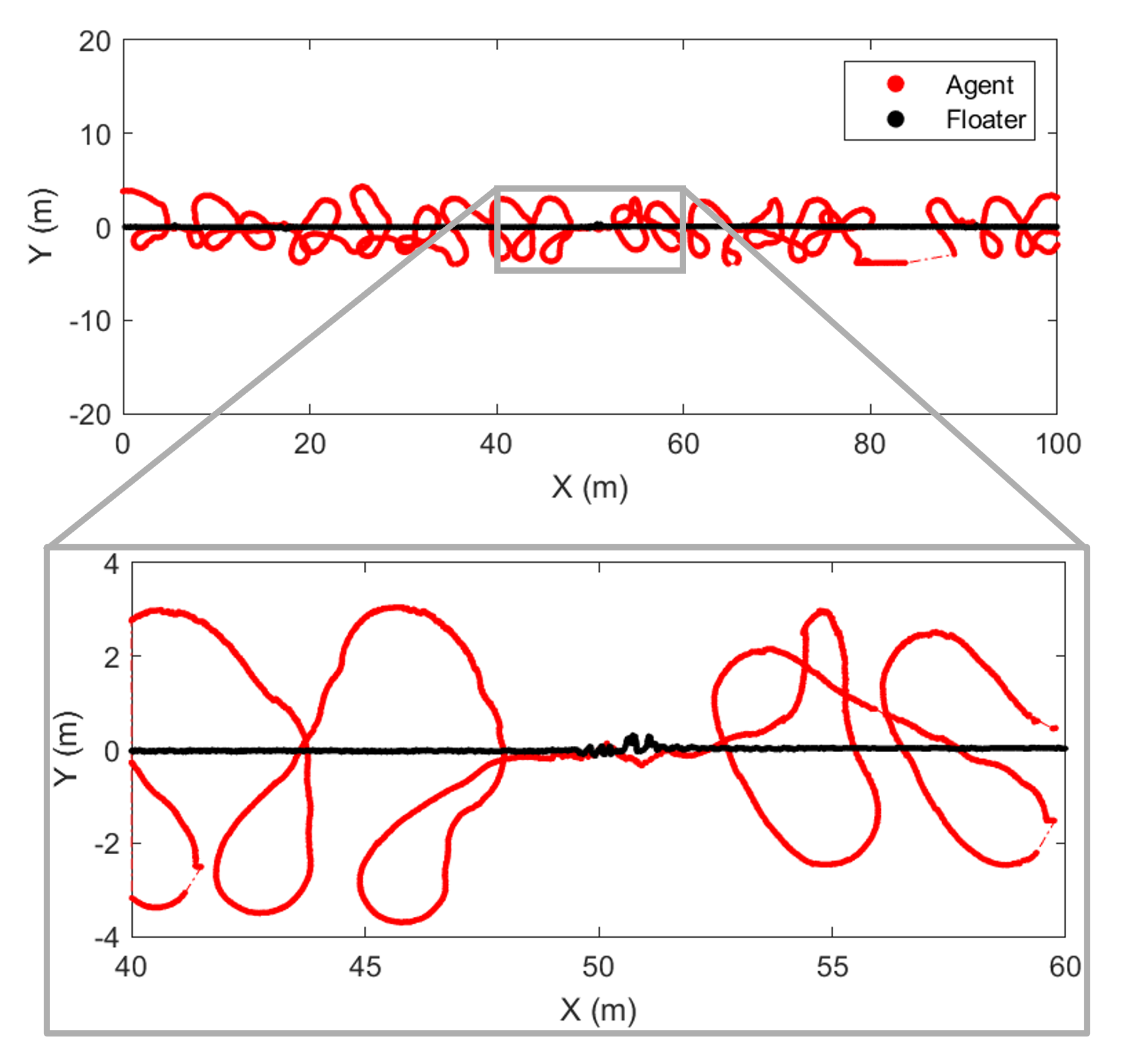}
    \caption{Extended Fencing path. Assuming moving floater moving at stream speed of 0.08 $m/s$. Full path above and close up below. }
    \label{Figure_16}
\end{figure}

The results obtained from the experiment can be "extended", in order to emulate Fencing of Agents around the Floater moving at a constant speed of 0.08 $m/s$. Considering the duration of the test of around 21 minutes, the Floater virtually covered about 100 $m$ of distance. Figure \ref{Figure_16} shows the extended path of Floater and Agent over the full test duration. The Fencing behavior on multiple Agents, with a moving Floater, can be utilized as a method to perform a schooling behavior. Quantitative results are showcased in Table \ref{Table_6}.

\subsection{Milling Pool Test}
\label{subsec:MilPool}
For testing the Milling behavior, the RVB algorithm was selected. We set a constant Agent speed of around 0.2 $m/s$ and a circle path with $R_0=$ 2 $m$ in a clockwise direction. Based on the Algorithm \ref{Algorithm_2}, the selected parameters values are: $k=20$ and $k_{rate}=2$. Tests were performed with one, two, and three Agents.

\begin{figure*}[!hbt]
    \centering
    \includegraphics[width=1\textwidth]{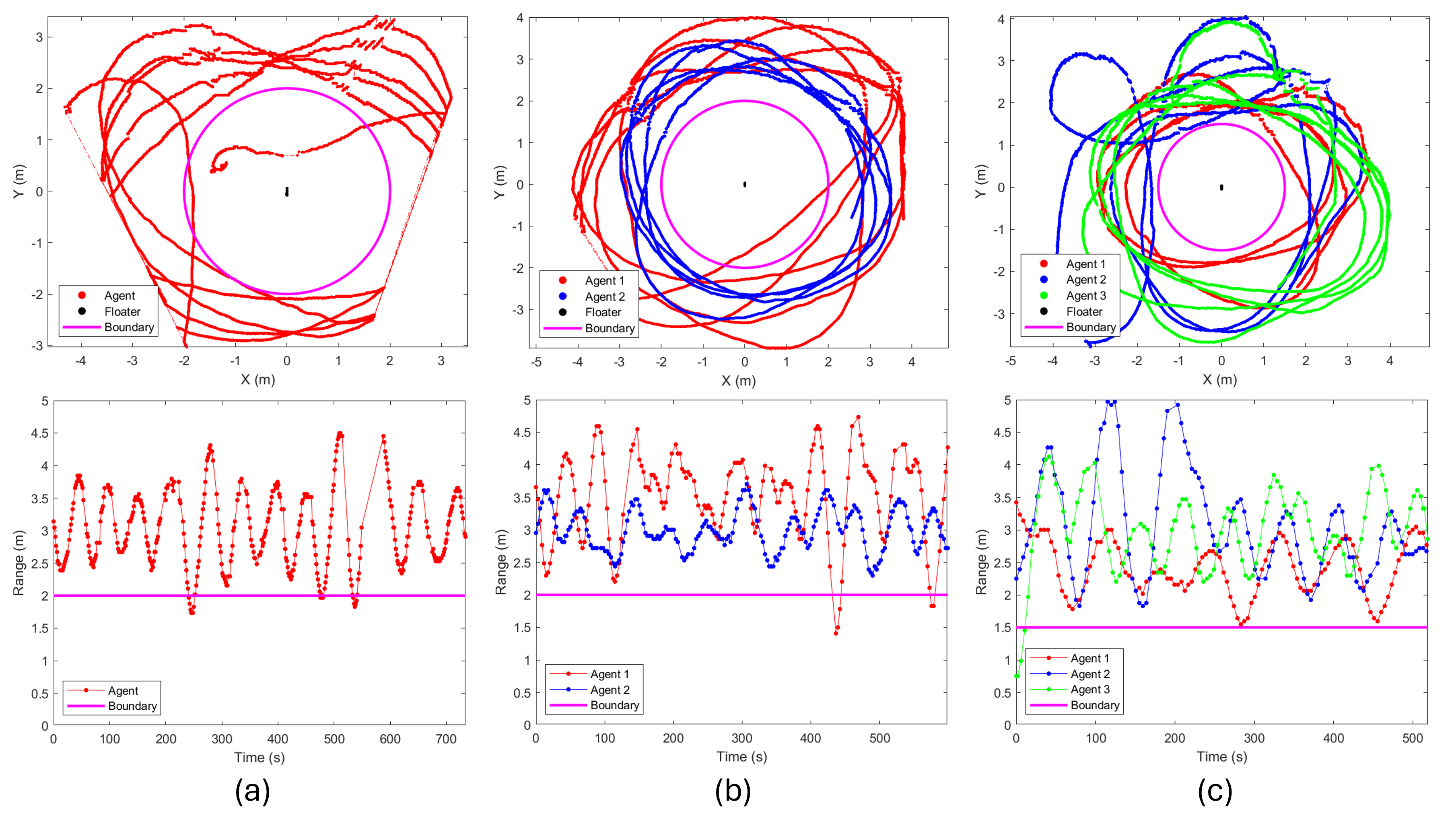}
    \caption{Milling results. Above, the reconstructed path, with desired paths in purple and the effective position of the floater in black. Below, are the relative acoustic ranges of Agents from Floater. a) Single Agent, b) Two Agents, c) Three Agents}
    \label{Figure_17}
\end{figure*}

Figure \ref{Figure_17} showcases paths and ranges of the Agents along the three tests. It is evident that the algorithm is effective at generating a clockwise rotatory motion around the beacon. It is valuable to remind that the Agents are not provided any other localization information other than the ranging information, for the entire duration of the tests.  

Ranging values showcase that the circular motion is at approximately 3 $m$ from the Floater while executing the Milling motion. This increase in radius, compared to $R_0$, is coherent with the effect observed in RVB Milling simulations (see Fig.\ref{Figure_7}(b)) as the effect of the measurement and communication delays. It is interesting to notice how Agent 2 in the three-Agents-test do temporarily inverts the direction of motion; this could have been caused by a multitude of factors but is most likely the consequence of a longer delay in range acquisition compared to the other tests. However, it is worth noting that the algorithm is able to autonomously return the Agent to the desired path and direction without any external intervention. Quantitative results are showcased in Tab.\ref{Table_7}.

\section{Field Experiments}
\label{Field Experiments}
\subsection{Experimental Site}
\begin{figure}[!hbt]
    \centering
    \includegraphics[width=0.4\textwidth]{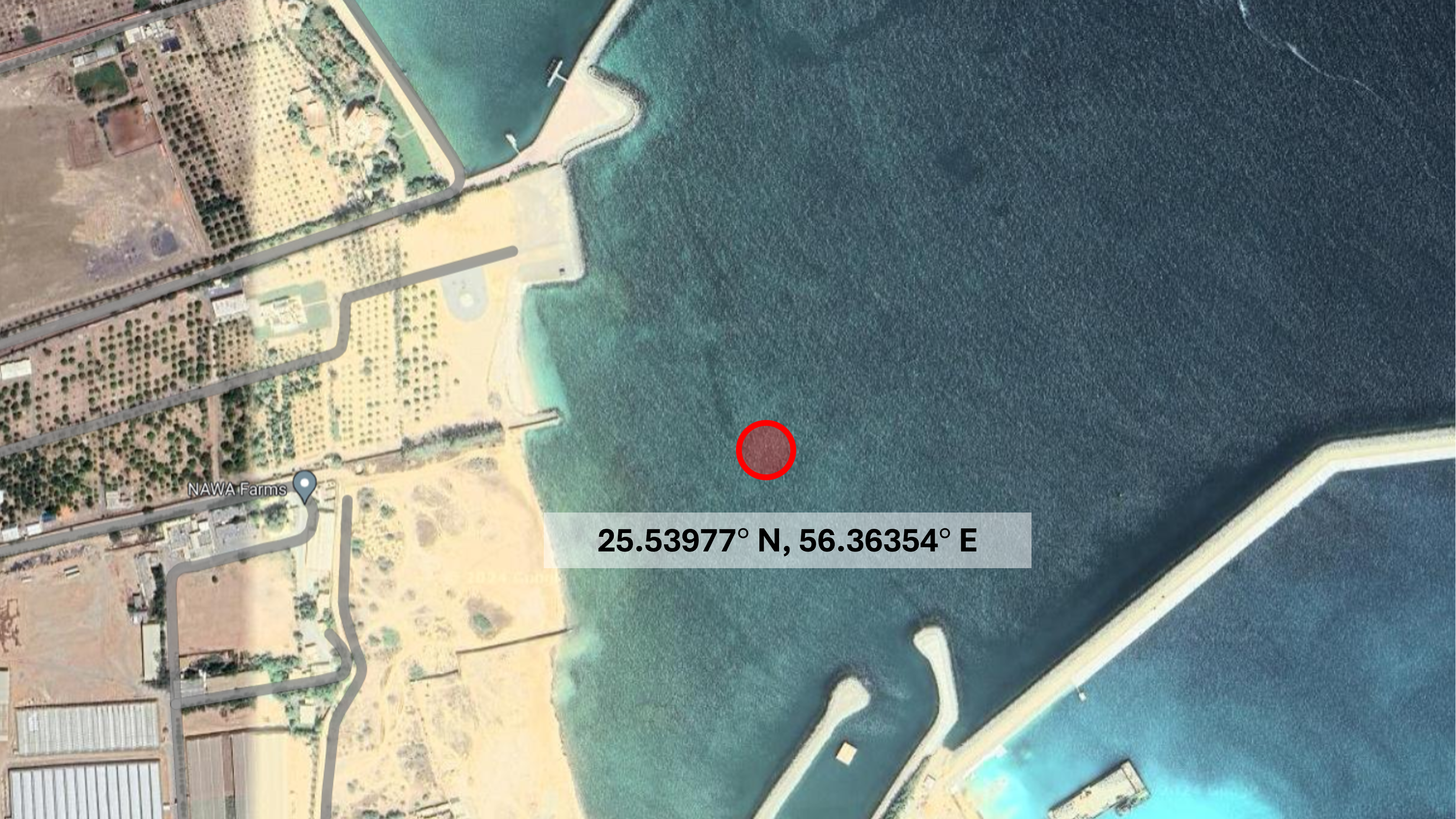}
    \caption{Location of Field Experiments [from Google Maps]. Al Fujairah Bay, UAE}
    \label{Figure_18}
\end{figure}

Field experiments were conducted in 07 June 2024 at Al Fujairah Bay in the United Arab Emirates. The location of the experiments is shown in Figure \ref{Figure_18}. The water depth in the location is around 2.5 $m$, with rocky and coral-rich seabed. During testing, we experienced weak underwater currents and small waves. The weather was sunny and the temperature was around 35$^\circ C$.

\subsection{Setup}
Fencing and Milling experiments were performed on a single Agent. A circle boundary (or path) with a 2 $m$ radius has been implemented in order to compare the experiments with the ones described in Section \ref{subsec:FenPool} and \ref{subsec:MilPool}. Similarly, the Agent's algorithm parameters matched the pool test conditions. For the Fencing experiment, the HEB model was activated, while for the Milling experiment, the RVB model was used. The Floater remained stationary, and the Agent maintained a constant depth, consistent with the pool tests. The main difference from the pool tests is an increase of actuation thrust $f_x$ to 1.0$N$, in order to counteract the added environmental disturbances. 

\subsection{Fencing Results}
\begin{figure}[!hbt]
    \centering
    \includegraphics[width=0.4\textwidth]{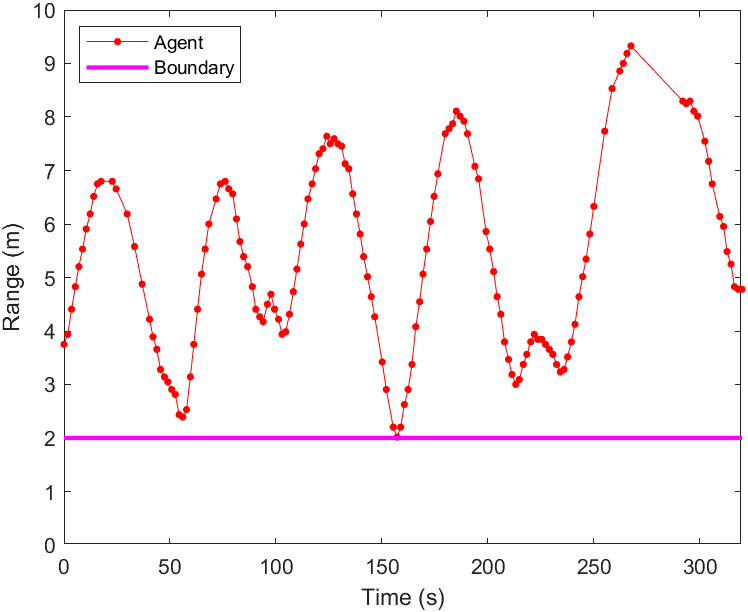}
    \caption{Agent ranges obtained from acoustic modem data during Fencing field test.}
    \label{Figure_19}
\end{figure}

The ranges of the Agent from the Floater are recorded during the Fencing experiment and are plotted in Figure \ref{Figure_19}. The experiment last approximately 320 $s$. It is clear that the Agent makes multiple attempts to come back towards the 2 $m$ radius circle. The maximum range reached from the floater is close to 9 $m$, which corresponds to an MRE of about 7 $m$.

The performance reduction is within expectations. An increase in $f_x$ proportionally affected the overshooting. Compounded it, waves and minor underwater currents affect both the platforms' movements and add uncertainty to the measurements. In particular, the assumption of constant speed is directly affected by the constant wave noise. Nevertheless, the Agent can roughly estimate the beacon direction and enact motion to close the relative distance on multiple instances without external intervention or other localization aids.

The mean range of the Agent during the Fencing experiment is 5.3 $m$. This indicates that the Agent was 3.3 $m$ away from the boundary line on average. The ART of the Agent is around 33 $s$. During testing, the Agent barely remains within the boundary. This is mostly due to the extremely small $R_0$ compared to the system dynamics. This is by design, both for the practicality of testing and to focus our study on the behavior outside the boundary. Finally, quantitative results are presented in Table \ref{Table_6}.

\subsection{Milling Results}
\begin{figure}[!hbt]
    \centering
    \includegraphics[width=0.4\textwidth]{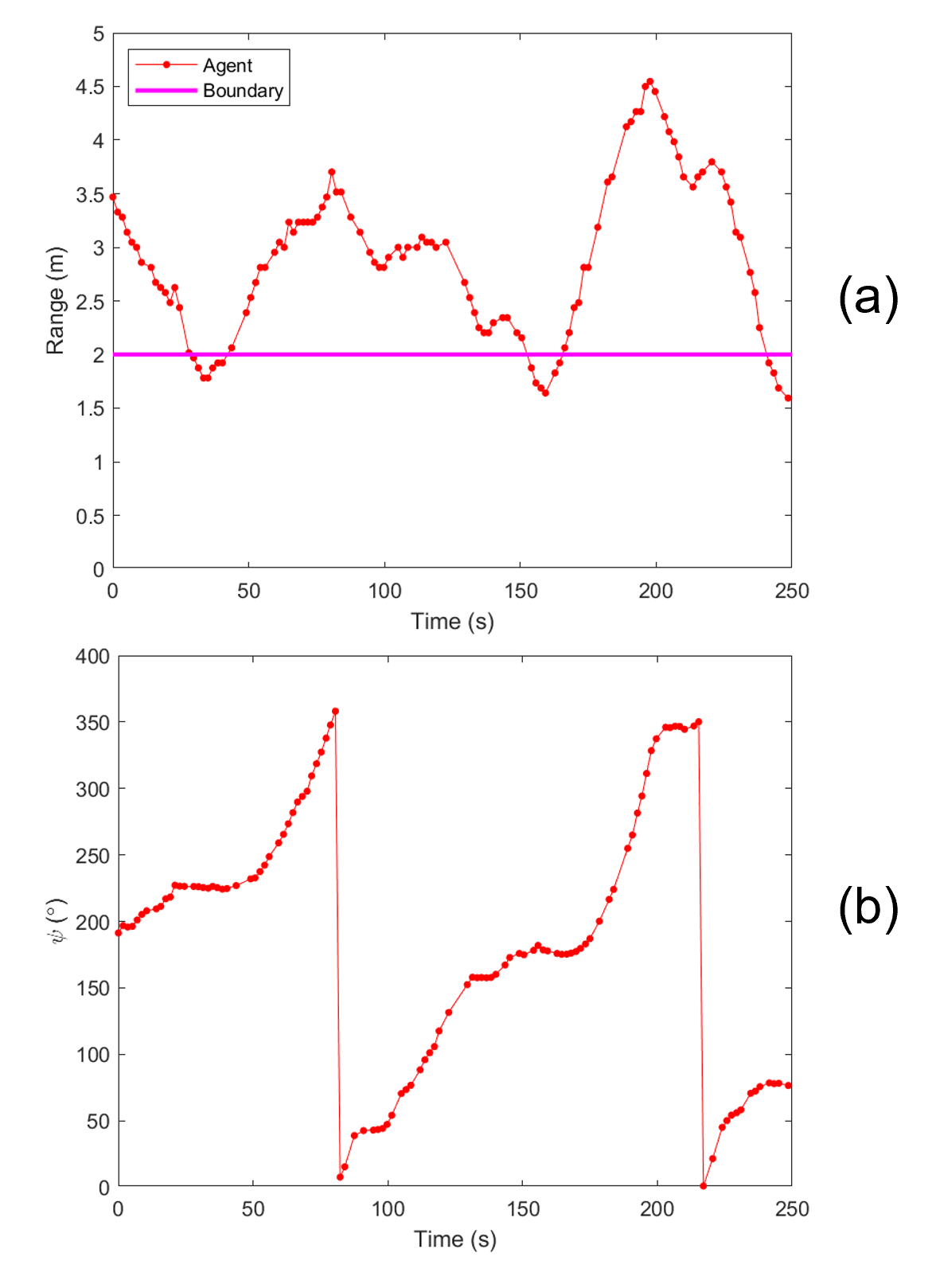}
    \caption{Sensor data obtained during Milling field test. (a) Ranges of Agent from Floater obtained through the acoustic modem, and (b) $\psi$ value of Agent obtained from IMU at each instance a range message is received}
    \label{Figure_20}
\end{figure}

The ranges of the Agent received through acoustic communication during the Milling experiment are shown in Figure \ref{Figure_20} (a). The average range $\overline{r}$ is 2.88 $m$, and the MRE reached is 2.5 $m$. This indicates a higher overshoot in the field test compared to the pool test. Similarly to the Fencing case, this can be attributed to the effect of higher $f_x$, underwater current, and wave disturbances. It is interesting to notice that the MRE is only 0.88 $m$, and the $\overline{\sigma}$ is 0.72 $m$, which is close to the results obtained during pool tests.

The heading values, collected from the IMU, are showcased in Figure \ref{Figure_20} (b) as proof of the rotatory motion of the Agent around the Floater. In a span of 250 $s$ of testing, the Agent is able to autonomously complete close to two full revolutions around the Floater. Quantitative results are presented in Tab.\ref{Table_7}.

\section{Discussion}
\label{Discussion}
Table \ref{Table_6} compares HEB Fencing results between Simulation, Pool Tests, and Field Experiments. It highlights the impact of changing the total numbers of Agents and surge value $f_x$ on the performance metrics, such as MRE, MPE, and ART. From the results in Table \ref{Table_6}, we can infer that the increase in the number of Agents cause an increment in all the recorded metrics values. This is due to the higher delay in receiving range messages for each Agent, and it is noticeable both in simulation and particularly in experimental scenarios. 
Unsurprisingly, simulations outperform real-world results, as they do not account for disturbances and sensor uncertainties. In particular, the ART results are the most affected, highlighting that Agents are noticeably less reactive to the Fencing Protocol. 
Nonetheless, by comparing quantitative and qualitative results, we can assess that the simulation can be used as a base for designing Fencing methods. Of particular interest are the Fencing results obtained by testing with current: MRE and MPE are worse than the static Fencing but with improved ART results. This outcome is expected because of the higher surge speed, which helped the Agent to go back to the boundary more rapidly. Overall, the results highlight the trade-offs between precision, speed, and number of Agents in a controlled environment.

In the field test, the parameters applied are the same as the pool test with underwater current, except for the circle boundary, whose radius was increased from 1.5 to 2 $m$. 
All of the metrics evaluated showcased a significant reduction in performance when exposed to a real-world environment. As discussed previously, among the main causes hypothesized, is the effect of unstructured underwater currents and wave disturbance, which particularly affected the assumption of constant speed on which the model is based. Furthermore, the algorithm is sensitive to errors in the rate-of-change of ranges, which is also affected by disturbances. Also, we detected an effective reduction of the rate of update (due to data loss in acoustic communication), with a reduction of update frequency from 0.625 $Hz$ on average in the pool to 0.455 $Hz$ during the field test. As we proved already with the addition of multiple Agents, update frequency directly affects performances, and the increase of data loss can probably be attributed to the shallow and irregular seabed. It is worth noticing that Field tests are performed at higher base speeds ($f_x$) in order to counteract disturbances. It is a reasonable assumption that even higher surge velocities would positively impact the velocity-to-disturbance ratio, improving the method's overall performance.

A final result to be highlighted, comparing simulated and experimental results, is that boundary size does not affect the presented results. As such, Fencing methods would be well suited for Agents confined to large boundaries and moving at sustained surge speeds.

\begin{table}[!ht]%
\centering
\caption{Comparison of HEB Fencing results between Simulation, Pool Tests, and Field Experiment  \label{Table_6}}
    \centering
\begin{tabular}{>{\centering\arraybackslash}m{0.2\columnwidth} >{\centering\arraybackslash}m{0.15\columnwidth} >{\centering\arraybackslash}m{0.05\columnwidth} >{\centering\arraybackslash}m{0.1\columnwidth} >{\centering\arraybackslash}m{0.1\columnwidth} >{\centering\arraybackslash}m{0.1\columnwidth}}
    \hline
        \textbf{Setting} & \textbf{Number of Agent}s & \textbf{$f_x$ (N)} & \textbf{MRE (m)} & \textbf{MPE (m)} & \textbf{ART (s)} \\ \hline
        ~ & 1 & 0.5 & 1.458 & 0.698 & 8.437 \\
        Simulation. (Circle 30$m$) & 2 & 0.5 & 1.461 & 0.706 & 10.113 \\
        ~ & 3 & 0.5 & 1.905 & 0.778 & 13.531 \\ \hline
        ~ & 1 (2$m$) & 0.5 & 1.891 & 1.135 & 30.910 \\
        Pool & 2 (2$m$) & 0.5 & 3.484 & 2.393 & 39.833 \\
        Test & 3 (1.5$m$)& 0.3 & 4.031 & 3.537 & 56.399 \\
        ~ & 1 (1.5$m$, current = 0.08 $m/s$) & 1 & 2.933 & 2.037 & 21.482 \\ \hline
        Field Test & 1 (2$m$) & 1 & 7.328 & 5.721 & 57.165 \\
    \hline
    \end{tabular}
\end{table}

Table \ref{Table_7} showcases the comparison of Milling results obtained from simulations, pool tests, and field tests. All simulation results on a 30 $m$ radius showcase extremely good precision $\overline{\sigma}$ ad accuracy $\overline{\mu}$, and it is worth noting that the performance of HEB and RVB are comparable. These results demonstrate the effectiveness of the Milling algorithms in the absence of noise. More interestingly, accuracy value $\overline{\mu}$ grows strongly when attempting much smaller $R_0=2 \ m$: This bias between target radius and effective radius can also be clearly observed in all pool testing. 

Regarding the pool testing, while the results in terms of precision are worse than the simulated, which is a consequence of disturbances and uncertainties of the sensors, accuracy results follow very closely the simulated results. Furthermore, we can notice that, while the increased number of Agents does negatively affect accuracy,  the phenomenon is not as strongly marked as for Fencing results. Finally, MRE values are consistent among all RVB tests.

As for the RVB Milling field test, the $\overline{\sigma}$ is higher than all of the pool tests, which means that there were high fluctuations in the ranges of the Agent. Surprisingly, the MRE and $\overline{\mu}$ results are better than in all other tests. This is despite range update rate reduction from 0.55 $Hz$ (single Agent pool test) to 0.45 $Hz$ in the field test. We hypothesize that a higher $f_x$ allowed the Agent to better compensate for the environment disturbances. Moreover, it also increased the absolute range-rate values, which resulted in an increased accuracy. Anyway, these results strengthen the assumption that RVB algorithms, while less flexible than HEB, are less affected by external disturbances.

\begin{table}[!ht]%
\centering
\caption{Comparison of Milling results between Simulation, Pool Tests, and Field Experiment  \label{Table_7}}
    \centering
\begin{tabular}{>{\centering\arraybackslash}m{0.2\columnwidth} >{\centering\arraybackslash}m{0.15\columnwidth} >{\centering\arraybackslash}m{0.05\columnwidth} >{\centering\arraybackslash}m{0.1\columnwidth} >{\centering\arraybackslash}m{0.1\columnwidth} >{\centering\arraybackslash}m{0.1\columnwidth}}
    \hline
        \textbf{Setting} & \textbf{Number of Agent}s & \textbf{$f_x$ (N)} & \textbf{MRE (m)} & \textbf{$\overline{\sigma}$ (m)} & \textbf{$\overline{\mu}$ (m)} \\ \hline
        ~ & 1 & 0.5 & 0 & 0.0177 & $-$0.160 \\
        Simulation (HEB Circle 30$m$)& 2 & 0.5 & 0 & 0.0397 & $-$0.102 \\
        ~ & 3 & 0.5 & 0.0879 & 0.0625 & 0.00501 \\ \hline
        Simulation & 3 (30$m$) & 0.5 & 2.875 & 0.00976 & 0.0970 \\
        (RVB) & 3 (2$m$) & 0.5 & 3.321 & 0.133 & 1.221 \\ \hline
        ~ & 1 (2$m$) & 0.5 & 2.498 & 0.598 & 1.021 \\
        Pool Test & 2 (2$m$) & 0.5 & 2.734 & 0.484 & 1.258 \\
        ~ & 3 (1.5$m$)& 0.3 & 3.469 & 0.617 & 1.314 \\ \hline
        Field Test & 1 (2$m$)& 1 & 2.551 & 0.717 & 0.883 \\
    \hline
    \end{tabular}
\end{table}

\section{Conclusions}\label{Conclusions}
This paper presents two distinct models for performing boundary control on multi-AUVs. The RVB model depends solely on range information obtained through acoustic communication, while the HEB model depends on ranges, range rate, and AUV heading. Fencing and Milling behaviors are tested in simulations, pool tests, and field tests. Results show the successful implementation of the behaviors, which validate the proposed models. Future work will include the implementation of direct Doppler velocities instead of the rate-of-change of consecutive ranges. This would offer significant scalability to the methods, as the updated rate of Doppler velocity is not affected by the number of AUVs, largely increasing the update rate to the HEB models, thereby making the method particularly appealing for multi-robots and swarm systems.

\section*{ACKNOWLEDGMENT}
This study was supported by the Technology Innovation Institute  (contract no. TII/ARRC/2047/2020) and by Khalifa University under Awards No. RC1-2018-KUCARS. Special thanks to Juan Florez, Carlos Parrilla and the Fujairah Research Center (FRC) for their support during field tests. Thanks to Jacob George and the KU-Marine lab for the experimental support.

\bmsection*{Data Availability Statement}

The data that support the findings of this study are available from the corresponding author upon reasonable request.

\bibliography{main}

\nocite{*}

\end{document}